\pdfoutput=1
\documentclass[preprint,12pt]{elsarticle}

\usepackage[utf8]{inputenc} 
\usepackage[T1]{fontenc}    
\usepackage{hyperref}       
\usepackage{url}            
\usepackage{booktabs}       
\usepackage{amsfonts}       
\usepackage{nicefrac}       
\usepackage{microtype}      
\usepackage{graphicx}
\usepackage{amsmath,amssymb}
\usepackage[table]{xcolor}
\usepackage{siunitx}        
\usepackage{placeins}       
\usepackage{multirow}
\usepackage{pifont}
\usepackage[english]{babel}
\definecolor{lightgreen}{rgb}{0.88, 1, 0.88}
\definecolor{lightred}{rgb}{1, 0.88, 0.88}
\newcommand{\yes}{\checkmark}
\newcommand{\no}{$\times$}

\usepackage{tabularx}       
\usepackage{threeparttable} 
\usepackage{array}          

\definecolor{sdu_red}{RGB}{147, 27, 49}
\hypersetup{
    colorlinks=true,
    linkcolor=sdu_red,
    filecolor=magenta,
    urlcolor=sdu_red,
    citecolor=blue,
}

\definecolor{colorCorrect}{RGB}{144, 238, 144} 
\definecolor{colorError}{RGB}{255, 99, 71}     

\definecolor{unetcolor}{RGB}{66, 133, 244}
\definecolor{deepcolor}{RGB}{234, 67, 53}
\definecolor{foundationcolor}{RGB}{52, 168, 83}
\definecolor{segformercolor}{RGB}{251, 188, 5}
\definecolor{legacycolor}{RGB}{103, 58, 183}

\newcommand{\segformer}{SegFormer}
\newcommand{\deeplab}{DeepLabV3+}
\newcommand{\medsam}{MedSAM}
\newcommand{\unet}{UNet}
\newcommand{\unetresnet}{UNet-ResNet34}
\newcommand{\hybrid}{MedSAM+UNet}
\newcommand{\sam}{SAM}
\newcommand{\fcn}{FCN}
\newcommand{\segnet}{SegNet}
\newcommand{\lightunet}{Lightweight U-Net}

\journal{Biomedical Signal Processing and Control}

\begin{document}

\begin{frontmatter}

\title{Challenges in Deep Learning-Based Small Organ Segmentation: A Benchmarking Perspective for Medical Research with Limited Datasets}


\author[sdu_cis]{Phongsakon Mark Konrad}
\ead{phon@mmmi.sdu.dk}

\author[sdu_cim]{Andrei-Alexandru Popa}
\ead{andrei@sdu.dk}

\author[sdu_cim]{Yaser Sabzehmeidani}
\ead{yasers@sdu.dk}

\author[duke]{Liang Zhong}
\ead{zhong.liang@duke-nus.edu.sg}

\author[duke]{Madhulika Tripathy} 
\ead{madhulika.tripathi@duke-nus.edu.sg}

\author[bcrs]{Andrei Constantinescu} 
\ead{andrei.constantinescu@umfcd.ro}

\author[nhcs]{Elisa A. Liehn}
\ead{liehn.elisaanamaria@nhcs.com.sg}

\author[sdu_cis]{Serkan Ayvaz\corref{cor1}}
\ead{seay@mmmi.sdu.dk}

\affiliation[sdu_cis]{organization={Centre for Industrial Software, University of Southern Denmark},
            addressline={Alsion 2},
            city={Sønderborg},
            postcode={6400},
            country={Denmark}}

\affiliation[sdu_cim]{organization={Centre for Industrial Mechanics, University of Southern Denmark},
            addressline={Alsion 2},
            city={Sønderborg},
            postcode={6400},
            country={Denmark}}
            
\affiliation[duke]{organization={Duke-NUS},
            addressline={8 College Road},
            city={Singapore},
            postcode={169857},
            country={Singapore}}

\affiliation[nhcs]{organization={National Heart Center Singapore},
            addressline={5 Hospital Dr},
            city={Singapore},
            postcode={169609},
            country={Singapore}}
            
\affiliation[bcrs]{organization={University of Medicine and Pharmacy Carol Davila Bucharest},
            addressline={Bulevardul Eroii Sanitari 8},
            city={Bucharest},
            postcode={050474},
            country={Romania}}

\cortext[cor1]{Corresponding author}

\begin{abstract}
Accurate segmentation of carotid artery structures in histopathological images is vital for cardiovascular disease research. This study systematically evaluates ten deep learning segmentation models including classical architectures, modern CNNs, a Vision Transformer, and foundation models, on a limited dataset of nine cardiovascular histology images. We conducted ablation studies on data augmentation, input resolution, and random seed stability to quantify sources of variance. Evaluation on an independent generalization dataset ($N=153$) under distribution shift reveals that foundation models maintain performance while classical architectures fail, and that rankings change substantially between in-distribution and out-of-distribution settings. Training on the second dataset at varying sample sizes reveals dataset-specific ranking hierarchies confirming that model rankings are not generalizable across datasets. Despite rigorous Bayesian hyperparameter optimization, model performance remains highly sensitive to data splits. The bootstrap analysis reveals substantially overlapping confidence intervals among top models, with differences driven more by statistical noise than algorithmic superiority. This instability exposes limitations of standard benchmarking in low-data clinical settings and challenges assumptions that performance rankings reflect clinical utility. 
We advocate for uncertainty-aware evaluation in low-data clinical research scenarios from two perspectives. First, the scenario is not niche and is rather widely spread; and second, it enables pursuing or discontinuing research tracks with limited datasets from incipient stages of observations

\end{abstract}

\begin{highlights}
\item Benchmarking deep learning models on small medical datasets is highly unstable.
\item Model rankings are dataset-specific and do not generalize across datasets.
\item Statistical analysis reveals no significant difference between top models.
\item Model rankings are sensitive to chosen cross-validation protocol and data splits.
\item We advocate for uncertainty-aware evaluation in low-data clinical research scenarios.
\end{highlights}


\end{frontmatter}


\section{Introduction}\label{sec:introduction}

Histological analysis is fundamental to cardiovascular pathology, providing detailed information for establishing diagnoses and tracking the morphological changes that define diseases like myocardial infarction and in-stent restenosis \cite{birare2024study,dinc2023post,matsuhiro2020histological}. 

However, ensuring reproducibility of results across different laboratories remains a critical challenge. Although guidelines exist for standardizing instrument calibration and operation, documentation of microscopy protocols and data analysis procedures remains insufficient, compromising experimental reproducibility.

In the cardiovascular field, debates continue over the optimal methodologies for quantifying functional and morphological changes following surgical procedure in murine models, like atherosclerosis or myocardial infarction. Key questions revolve around how plaque size or infarct size should be measured, how it should be expressed (as a percentage based on a single section, averaged across multiple sections, and should focus solely on affected regions or include the entire organ). The absence of consensus has led to a heterogeneous body of published data, which complicates the translation of findings and impedes the development of standardized measurement protocols. As a result, researchers must often reconstruct methodologies independently, slowing progress and reducing reproducibility in cardiovascular research

Computer-based systems and artificial intelligence (AI) can improve standardization and consistency in cardiovascular research. While these tools are typically applied to large datasets, smaller datasets from animal studies receive less attention. The cardiovascular community increasingly seeks to minimize animal use \cite{hubrecht20193rs} and obtain maximum information from minimal samples, even though current AI applications like 3D reconstructions do not fully support this approach. The variety of methods currently used raises concerns about the reliability of tissue analysis results. Therefore, efforts are being made to compare different methods on the same limited datasets to improve consistency and standardization across laboratories, ultimately strengthening the reproducibility of cardiovascular research.

As an example, the detailed characterization of vascular lesions from histological sections is especially critical in understanding conditions like carotid artery stenosis, a primary contributor to ischemic stroke \cite{arasu2021carotid}. However, the translation of this vital technique from research to widespread clinical application is present with challenges. The current gold standard, which relies on manual analysis of tissue by trained personnel, that is labor-intensive \cite{roberts2020diagnostic}. This manual process not only risks damaging the fragile tissue samples but it introduces significant inter-observer variability, a fundamental barrier to producing reproducible science \cite{pandelea2020surface}.

In recent years, deep learning models, from established Convolutional Neural Networks (CNNs) to massive, pre-trained Foundation Models (FMs), have shown promising results \cite{ahmad2020deep, de2021machine, kong2020toward, shamai2022deep}. However, these models are typically developed on large datasets, whereas specialized medical research often operates under data-scarce conditions \cite{basla2024expert}. This raises a fundamental methodological concern: when only a handful of patient samples are available, standard methods for selecting the ``best'' model may themselves be unreliable.

This study systematically investigates the performance of state-of-the-art deep learning models for carotid artery segmentation in the context of limited cardiovascular histopathological data. Our objective was to identify effective model architectures for this data-constrained task. We conducted a benchmarking study involving ten architectures: classical CNNs (FCN, SegNet), modern encoder-decoder networks (U-Net variants \cite{ronneberger2015unet}, DeepLabV3+ \cite{chen2018encoder}), a lightweight transformer (SegFormer \cite{xie2021segformer}), and foundation models (SAM \cite{kirillov2023sam}, MedSAM \cite{ma2024medsam}, MedSAM+UNet \cite{yang2024samunet}). Beyond performance comparison, we conducted systematic ablation studies examining data augmentation strategies (100 experiments across 10 presets), input resolution sensitivity (128--1024 pixels), and random seed stability (5 seeds per model) to quantify multiple sources of experimental variance. 

The evaluation methodology followed widely adopted practices in machine learning research, incorporating  hyperparameter tuning via a thorough Bayesian optimization strategy and multiple cross-validation schemes \cite{cooper2021hyperparameter}. While such approaches aim to reduce variability and enhance reliability, they are not immune to the well-documented "illusion of control", a bias that can lead researchers to overestimate the stability of model performance in low-data regimes \cite{langer1975illusion}. 

Beyond model selection, this study tests the hypothesis that standard benchmarking leaderboards are unreliable in low-data regimes, and that observed model rankings may reflect statistical noise and evaluation choices rather than true algorithmic differences. Using a dataset of $N=9$ images as a test case, we examine the limitations of common benchmarking practices when data is scarce, and propose an evaluation framework centered on statistical equivalence and practical efficiency

\section{Related Work}
\label{sec:related_work}
\subsection{CNN-based Architectures for Medical Segmentation}
Convolutional Neural Networks (CNNs) have dominated medical image segmentation for over a decade. The U-Net architecture \cite{ronneberger2015unet} remains the most widely adopted model, with numerous variants such as UNet++ \cite{zhou2020unet++} and R2U-Net \cite{alom2019recurrent} extending its encoder-decoder design. As noted by \cite{paheding2024unetanalysis}, these models have facilitated the translation of AI research to clinical application. DeepLabV3+ \cite{chen2018encoder}, with its Atrous Spatial Pyramid Pooling (ASPP) module for multi-scale feature extraction, provides an alternative approach. However, the performance of both U-Net and DeepLabV3+ remains fundamentally linked to the availability of large annotated datasets \cite{yu2021convolutional}, a condition often unmet in specialized clinical research.

\subsection{Transformer-based Architectures in Vision and Medicine}
Transformer-based architectures leverage self-attention to model long-range dependencies, addressing the local receptive field limitations of CNNs \cite{vaswani2017attention}. SegFormer \cite{xie2021segformer} combines a hierarchical Transformer encoder with a lightweight MLP decoder, offering an efficient alternative to CNN-based segmentation. While such architectures provide broader representational capacity \cite{chrysos2024architecture}, they also introduce new validation challenges \cite{xia2023recent}.

\subsection{Foundation Models and Their Adaptation for Medical Imaging}
The Segment Anything Model (SAM) \cite{kirillov2023sam}, trained on over a billion masks, represents an important advance in general-purpose segmentation. Adapting SAM for medical imaging has been explored through domain-specific fine-tuning, as demonstrated by MedSAM \cite{ma2024medsam}, and through hybrid architectures that combine foundation model encoders with CNN-based decoders \cite{yang2024samunet}. However, the out-of-the-box performance of SAM on fine-grained medical tasks remains limited, necessitating careful adaptation strategies.

\subsection{Hyperparameter Optimization in Medical AI}
Model performance varies substantially with hyperparameter choices including learning rate, optimizer, scheduler, and loss function \cite{lai2024application}. The resulting high-dimensional search space makes exhaustive grid search infeasible, and random search provides no optimality guarantees \cite{subacsi2024comprehensive}. Bayesian optimization addresses this by building probabilistic surrogate models to guide the search efficiently \cite{agrawal2024survey}. The challenge is compounded when Parameter-Efficient Fine-Tuning (PEFT) techniques such as LoRA \cite{hu2022lora} introduce additional hyperparameters that must themselves be tuned.

\subsection{The Benchmarking Reliability on Small Medical Datasets}
A broader methodological concern arises when comparing models in the low-data regimes common in medical research \cite{salehi2023study, hong2023overcoming, maier2018rankings}. The practice of declaring a state-of-the-art model based on benchmark rankings is increasingly scrutinized \cite{lin2018neural}, as a model's rank can depend heavily on the evaluation protocol and data split \cite{lin2021significant}. LOOCV, while nearly unbiased, suffers from high variance \cite{ranglani2024empirical}, and extensive hyperparameter optimization risks overfitting to the validation set \cite{makarova2021overfitting, hertel2022reproducible}, producing results that may not reproduce on new data \cite{tetko2024aware}. Few studies address the stability of the evaluation process itself; our work addresses this gap by empirically quantifying how these concerns manifest in medical image segmentation.

\section{A Systematic Model Evaluation Framework}

\begin{figure}[h!]
\centering
\includegraphics[width=1\textwidth]{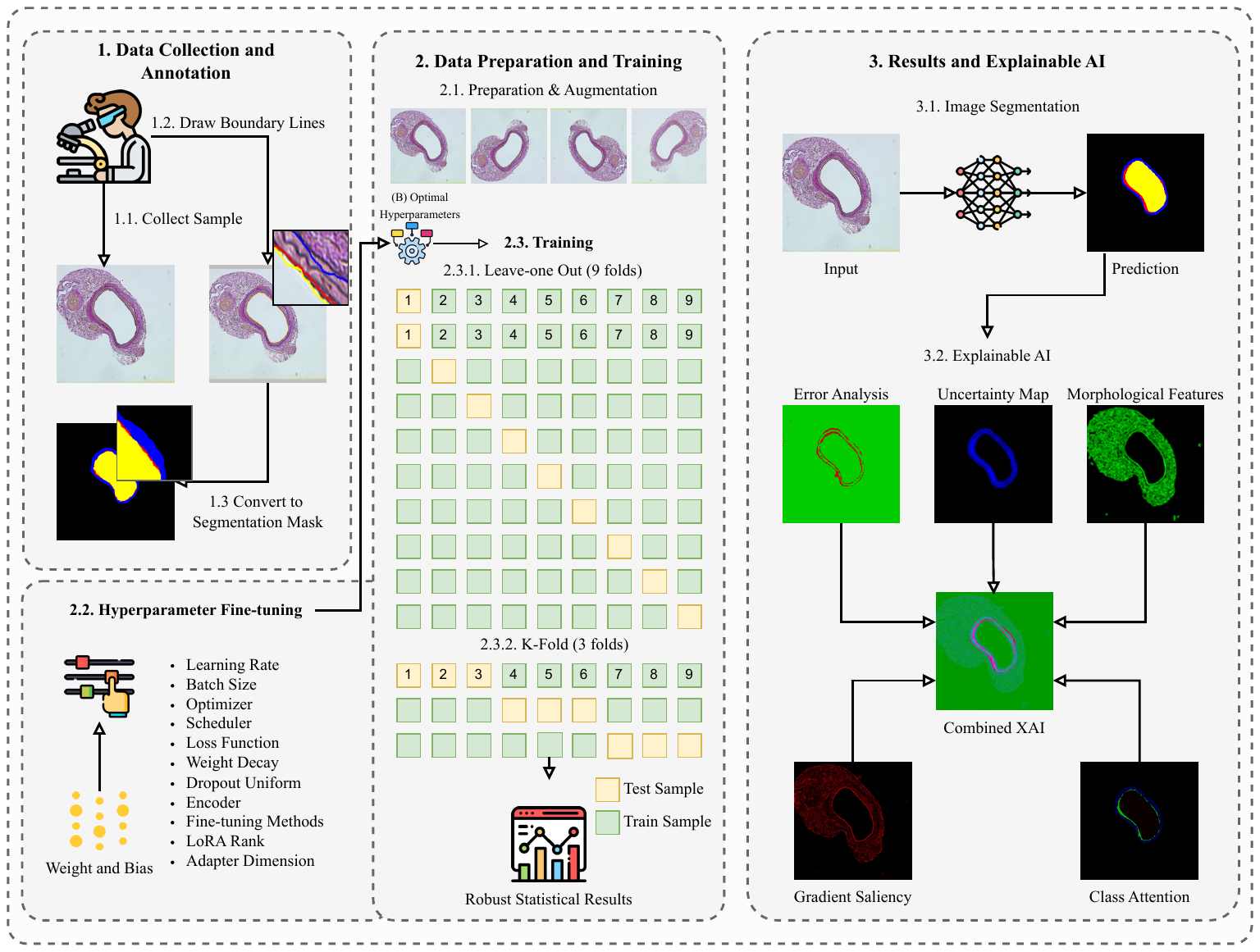}
\caption{The systematic evaluation framework. Our process begins with (1) data collection and expert annotation. After the data preparation (2.1) Each model architecture then undergoes (2.2) extensive Bayesian hyperparameter optimization to find its optimal configuration and then use it for the training (2.3). (3.1) We then evaluate the segmentation performances, which are then dissected using our (3.2.) multi-modal XAI framework for qualitative insight.}
\label{fig:framework_pipeline}
\end{figure}

To investigate the stability of deep learning benchmarks on small medical datasets, we designed a multi-stage evaluation framework. This process, illustrated in Figure \ref{fig:framework_pipeline}, was developed to ensure a fair and comprehensive comparison across diverse model architectures by systematically controlling for common sources of variability.

The pipeline begins with data collection and expert annotation (1). Rather than using arbitrary model configurations, each architecture undergoes systematic hyperparameter optimization (2.2.)---either Bayesian search or manual tuning based on established practices---to ensure near-optimal performance. This step is necessary for mitigating the confounding effect of suboptimal configurations when comparing models \cite{agrawal2024survey, lai2024application}. To generate statistically sound results, we then subject each optimized model to multi-strategy cross-validation (2.3.1, 2.3.2) in combination with on-the-fly image augmentation (3.1), employing both Leave-One-Out (LOOCV) and 3-Fold splits. LOOCV, in particular, has been shown to provide robust statistical estimates for small datasets \cite{xu2023small}. The quantitative results (4.1) are then dissected using our five-layer Explainable AI (XAI) framework (4.2) to identify potential sources of performance instability across data splits \cite{patil2024explainable, islam2024unified}.

\subsection{Dataset and Clinical Context}

\begin{figure}[h!]
\centering
\includegraphics[width=0.8\textwidth]{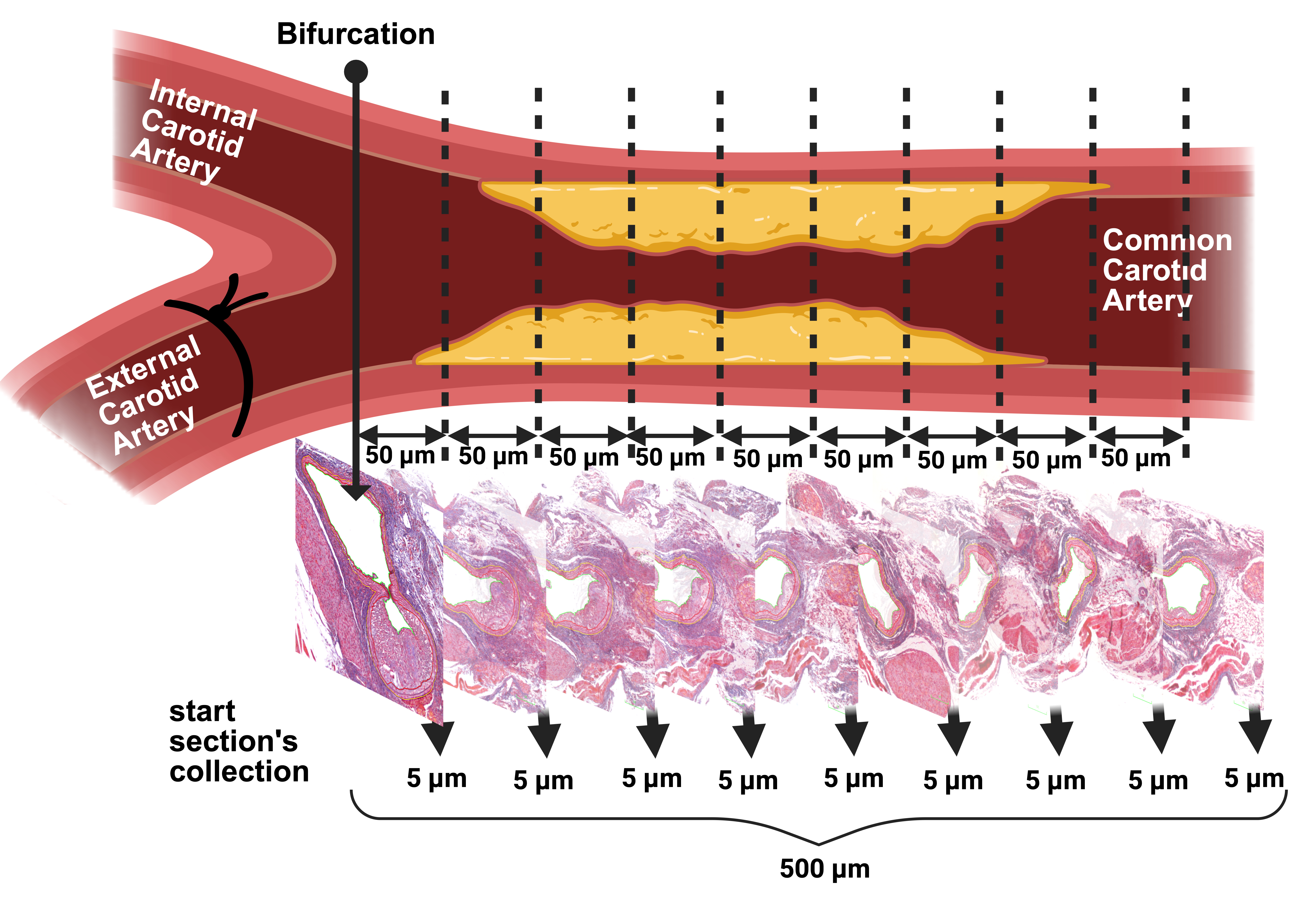}
\caption{5 µm serial sections were collected starting from the bifurcation from the atherosclerotic injured common carotid artery, 50 µm apart, as represented. Created in BioRender. Liehn, E. (2026) \url{https://BioRender.com/sm3xslo}.}
\label{fig:3d_reconstruction}
\end{figure}

The primary dataset (DS1) consists of nine high-resolution histological images of vascular tissue from the injured carotid artery of mice. Mice (C57/Bl6, LDL knockout male) undergo the carotid de-endothelialization, as described before by our group \cite{curaj2020induction} (Permission no.231805 from IACUC at the Biology Resource Center at the Agency for Science, Technology and Research (A*STAR), Singapore). The carotid artery was excised 2 weeks after de-endothelialization and embedded in paraffin. Serial sections (5 µm thick) were collected starting with the bifurcation between the internal carotid artery and external carotid artery. 10 sections, 5 µm thick and 50 µm apart were stained with orcein and photographed (Figure \ref{fig:3d_reconstruction}). These source images are stored in BMP format with variable native resolutions (e.g., $984 \times 792$ pixels). 

This small sample size ($N=9$) was intentionally chosen to reflect a clinical research scenario where data is scare, requiring model robustness and data efficiency. The task is a four-class semantic segmentation problem designed to delineate key vascular structures: Lumen (class 1), Neointima (class 2), and Media (class 3), against the Background (class 0). 

Ground truth annotations were independently performed by two domain experts (E.A. Liehn and A. Constantinescu) following the established annotation protocol \cite{curaj2020induction}, which defines the delineation of lumen, neointima (intima), and media boundaries using the lamina interna and lamina externa as reference landmarks. The two annotators reached full agreement through the structured protocol; initial discrepancies were resolved by consensus before finalizing the ground truth. While a formal inter-annotator agreement score (e.g., inter-rater Dice) was not computed, the use of well-defined anatomical landmarks (lamina interna and lamina externa) as reference boundaries minimizes subjective variability, as demonstrated in prior work using this protocol \cite{curaj2020induction}. 

\begin{figure}[h!]
\centering
\includegraphics[width=1\textwidth]{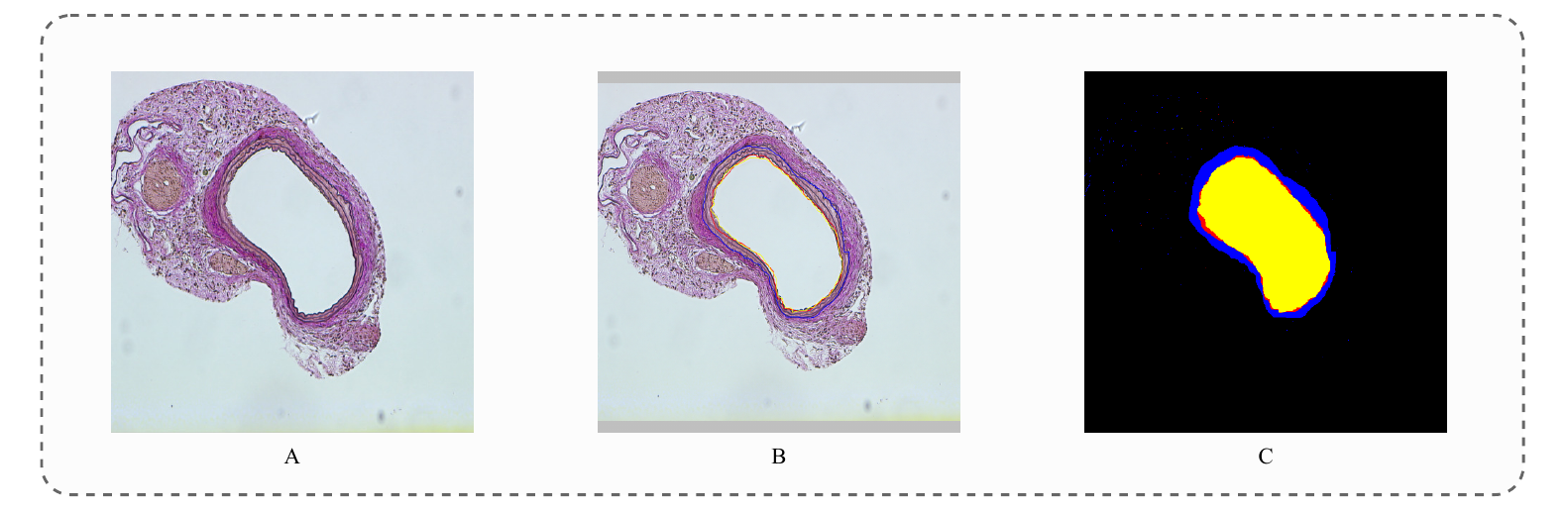}
\caption{Example of data preparation. (A) Original histological image. (B) Expert line-art annotation. (C) Processed ground truth segmentation mask for \textcolor{yellow!90!black}{Lumen} (yellow), \textcolor{red}{Neointima} (red), and \textcolor{blue}{Media} (blue).}
\label{fig:dataset_example}
\end{figure}

Ground truth masks were generated from these expert annotations; initial color-coded JPG annotations were processed into four-class, integer-labeled PNG masks for model training and evaluation. For use in our pipeline, these high-resolution source images and masks are systematically preprocessed, including resizing to standardized input dimensions (e.g., $256 \times 256$ pixels for conventional models). An example is shown in Figure \ref{fig:dataset_example}.

An independent generalization dataset (DS2) was collected from a separate cohort of LDLR knockout mice (C57/Bl6 background) with carotid de-endothelialization performed as described above. 
Serial sections (5 µm thick) from 22 carotid arteries were stained with orcein and photographed, yielding a total of $N=153$ images in TIF format. Three images were held out as a fixed independent test set; the remaining 150 were used for training. This dataset was used for two complementary experiments: (1) out-of-distribution generalization testing, where models trained on DS1 were evaluated directly on DS2 (Section \ref{sec:generalization}); and (2) in-distribution training at varying sample sizes ($N=9, 25, 50, 100, 150$), where models were trained on nested DS2 subsets and evaluated on the three held-out DS2 images, to assess dataset-specific learning curves and ranking stability (Section \ref{sec:ds2_indistribution}). DS2 was not used during DS1-based hyperparameter selection. Ground truth annotations followed the same protocol as DS1, with segmentation of lumen, neointima, and media structures.

\subsection{Data Preprocessing and Augmentation}

Our data preparation strategy employed a unified pipeline for all model architectures to ensure a fair comparison. All images and their corresponding masks were resized to a uniform dimension of $256 \times 256$ pixels and normalized to the range $[0, 1]$. For the foundation models (SAM and MedSAM), whose Vision Transformer encoders (ViT-B, ViT-L) were originally designed for $1024 \times 1024$ inputs, the upscaling to the encoder's native resolution is handled internally within each model's forward pass via bilinear interpolation. The output is subsequently downsampled back to the input resolution for loss computation. This design ensures that all models receive identical input from the data pipeline, ensuring consistent preprocessing across architectures.

Following this unified resizing, we employed an on-the-fly data augmentation strategy during training for all models to expand the limited dataset and improve generalization \cite{liu2024segmenting, xu2020automatic, dasari2022increasing}. This technique applied a series of random transformations to each image-mask pair at every epoch, encompassing both geometric and color-based changes. To account for variations in sample orientation, geometric augmentations included random horizontal and vertical flips, alongside affine transformations such as rotations ($\pm 20^\circ$), scaling (from 80\% to 120\%), and shearing. These transforms were applied synchronously to both the image and its segmentation mask to maintain perfect spatial alignment. Color augmentations (jittering of brightness, contrast, saturation, and hue) were applied exclusively to input images to simulate variations in staining and lighting conditions.

To ensure a unbiased evaluation, the validation and test sets for all models were subjected only to their respective resizing and normalization steps, without any data augmentation.

\subsection{Model Architecture Selection}
We selected ten models spanning three architectural paradigms (classical and modern CNNs, a Vision Transformer, and Foundation Models) to provide a diverse testbed for evaluating benchmark stability.

\subsubsection{Classical CNN Architectures}
\fcn{} (134.4M parameters) uses a VGG-16 backbone with ImageNet-pretrained weights and serves as a baseline for dense prediction. \segnet{} (29.5M parameters) employs an encoder-decoder design that reuses max-pooling indices during upsampling. Both serve as historical baselines.

\subsubsection{Modern CNN Architectures}
The standard \unet{} \cite{ronneberger2015unet} (31.4M parameters), trained from scratch, serves as the canonical encoder-decoder baseline. To evaluate transfer learning, we compared eight pre-trained encoders from the ResNet \cite{he2016resnet}, EfficientNet \cite{tan2019efficientnet}, DenseNet \cite{huang2017densenet}, and MobileNet \cite{howard2017mobilenets} families; the best-performing configuration was UNet-ResNet34 (24.4M parameters). A \lightunet{} variant (7.8M parameters) with depthwise separable convolutions tests whether reduced model capacity affects performance. \deeplab{} (26.7M parameters) captures multi-scale context through its ASPP module.

\subsubsection{Modern Vision Transformer}
SegFormer employs a hierarchical Transformer encoder with a lightweight MLP decoder, providing a computationally efficient alternative for semantic segmentation (3.7M parameters).

\subsubsection{Foundation Models and Hybrid Architectures}
We included SAM and its medical derivative MedSAM to assess the impact of large-scale pre-training. Since SAM requires manual prompts for segmentation, which is impractical for automated benchmarking \cite{khalili2024automatic, xu2024generalisability}, we implemented a learned auto-prompting module: a lightweight network uses multi-head cross-attention to fuse SAM's image embeddings with learnable prompt tokens, enabling fully automated end-to-end training. MedSAM was fine-tuned using Norm Tuning, a parameter-efficient strategy that updates only the normalization layers of the frozen encoder alongside the mask decoder. We also included a MedSAM-UNet hybrid that feeds the frozen MedSAM encoder features into a U-Net-style decoder, combining the context modeling of a Transformer with the spatial precision of a CNN decoder.

\subsection{Implementation Reproducibility}
\label{sec:reproducibility}

To ensure reproducibility and enable fair comparison across experimental conditions, we document the complete computational environment. All experiments were conducted using NVIDIA Tesla V100 GPUs (32GB VRAM) with PyTorch 2.1+ and CUDA 12.1. Automatic Mixed Precision (AMP) training was enabled to improve computational efficiency.

\subsubsection{Randomness Control}
Our implementation sets seeds across all relevant random number generators using a unified seed value ($seed = 42$), including Python's random module, NumPy's random state, and PyTorch CPU/CUDA operations. To ensure deterministic behavior, we configured PyTorch's cuDNN backend with deterministic mode enabled and benchmark mode disabled.

\subsubsection{Limitations of Single-Seed Evaluation}
A methodological limitation is the use of a single random seed for all experiments. While this ensures reproducibility of our specific results, it does not capture the variance from different random initializations. This limitation reinforces our central finding: even with controlled randomness, the small dataset size introduces substantial variance through the data splits themselves.

\subsection{Hyperparameter Optimization}

\begin{figure}[h!]
\centering
\includegraphics[width=1\textwidth]{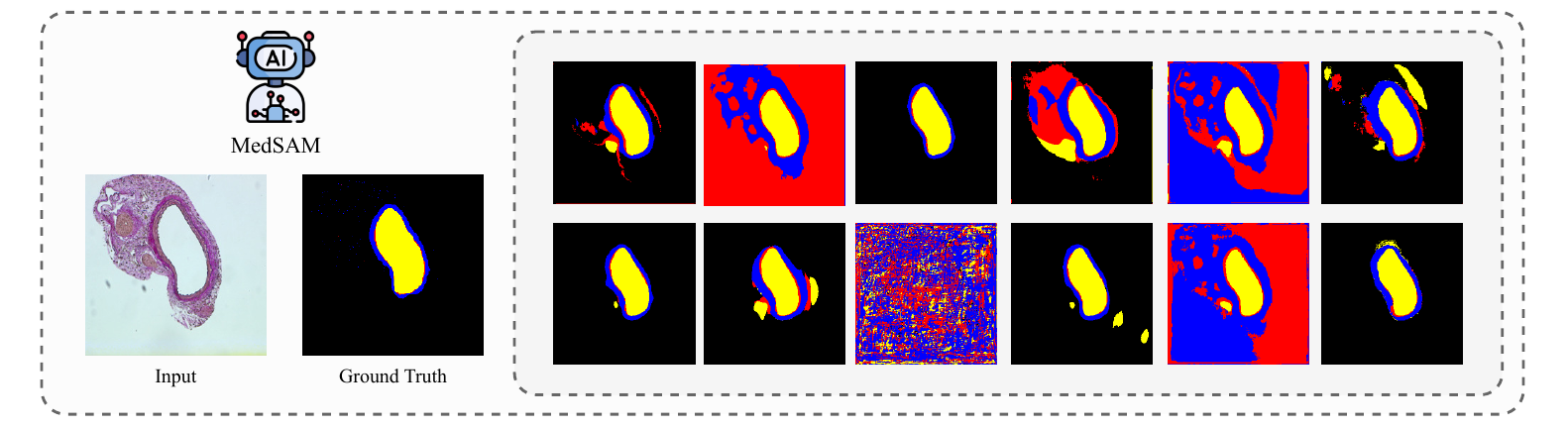}
\caption{The impact of hyperparameter selection. All twelve segmentation results were generated by the same MedSAM architecture. The wide variation in output quality is due solely to different hyperparameter configurations. This illustrates that without a rigorous search, one could incorrectly dismiss a capable model based on a single, suboptimal trial.}
\label{fig:dataset_example_before_sweep_supp}
\end{figure}

Hyperparameters significantly impact model performance, particularly with limited data \cite{lai2024application}. To account for this, we used Bayesian optimization (explained in \ref{sec:supp_bayesian}) managed by Optuna\footnote{https://optuna.org/} \cite{bergstra2011tpe}, a Bayesian hyperparameter optimization framework using the Tree-structured Parzen Estimator (TPE) sampler. Each of the ten architectures was independently optimized over 100 Bayesian trials, totaling 1,000 runs. Our extensive search space is detailed in Table \ref{tab:hyperparams}. As Figure \ref{fig:dataset_example_before_sweep_supp}  illustrates, this approach is crucial: without it, a strong architecture might be unfairly dismissed due to a single suboptimal configuration \cite{wang2023bayesian}. This process ensures each architecture is compared at its near-optimal configuration, which is detailed in Table \ref{tab:optimal_hyperparameters_styled}.

\begin{table}[h!]
\caption{Hyperparameter Search Space Summary.}
\label{tab:hyperparams}
\centering
\begin{tabularx}{\textwidth}{@{} l l X @{}}
\toprule
\textbf{Parameter} & \textbf{Search Strategy} & \textbf{Range / Values} \\
\midrule
\addlinespace[0.3em]
\multicolumn{3}{l}{\textit{\textbf{General Parameters}}} \\
\addlinespace[0.3em]
\midrule
Learning Rate & Log-uniform & [1e-5, 1e-2] \\
Batch Size & Categorical & [2, 4, 8, 16, 32] \\
Optimizer & Categorical & [Adam, AdamW, SGD, RMSprop, Nadam] \\
Scheduler & Categorical & [Cosine, Polynomial, Step, Warmup Cosine, OneCycle] \\
Loss Function & Categorical & [Focal-Dice, Focal-Tversky, Unified Focal] \\
Weight Decay & Log-uniform & [1e-6, 1e-2] \\
Dropout & Uniform & [0.0, 0.3] \\
\midrule
\addlinespace[0.3em]
\multicolumn{3}{l}{\textit{\textbf{Architecture-Specific Parameters}}} \\
\addlinespace[0.3em]
\midrule
Encoder (UNet) & Categorical & [ResNet18/\allowbreak 34/\allowbreak 50, EfficientNet-B0/\allowbreak B1/\allowbreak B2, DenseNet121, MobileNetV2] \\
SAM Model Type & Categorical & [ViT-B, ViT-L] \\
Fine-tuning Method & Categorical & [SVD, LoRA, Norm Tuning] \\
LoRA Rank & Categorical & [8, 16, 32] \\
Adapter Dimension & Categorical & [128, 256, 512] \\
\bottomrule
\end{tabularx}
\end{table}

\begin{table}[h!]
\begin{threeparttable}
\caption{Optimal Hyperparameters for Each Model. All ten models were independently optimized via Bayesian hyperparameter search (100 trials each). All models use Focal-Dice loss and train for 200 epochs.}
\label{tab:optimal_hyperparameters_styled}
\centering
\tiny
\setlength{\tabcolsep}{1pt}
\sisetup{detect-all}
\begin{tabularx}{\textwidth}{@{} l *{10}{>{\centering\arraybackslash}X} @{}}
\toprule
& \textbf{FCN} & \textbf{SegNet} & \textbf{LW-U}\tnote{d} & \textbf{UNet} & \textbf{U-R34}\tnote{c} & \textbf{DLV3+}\tnote{a} & \textbf{SF}\tnote{b} & \textbf{SAM} & \textbf{MS}\tnote{h} & \textbf{M-U}\tnote{e} \\
\midrule
\addlinespace[0.3em]
\multicolumn{11}{l}{\textit{Architecture}} \\
\addlinespace[0.2em]
Backbone           & VGG-16        & VGG          & Depth-sep     & Standard      & RN34\tnote{f}   & RN50          & MiT-B0        & ViT-B         & ViT-L         & ViT-B+U \\
Pretrained         & IN\tnote{f}   & ---          & ---           & ---           & IN              & IN            & IN            & SAM           & SAM           & SAM              \\
\addlinespace[0.3em]
\midrule
\addlinespace[0.3em]
\multicolumn{11}{l}{\textit{Optimization}} \\
\addlinespace[0.2em]
LR                 & 1e-4          & 1e-4         & 1e-4          & 1e-4          & 1e-4            & 5e-5          & 6e-5          & 6e-4          & 8.2e-4        & 8.7e-4           \\
Optimizer          & AdamW         & AdamW        & AdamW         & AdamW         & Adam            & Adam          & Adam          & AdamW         & Adam          & Adam             \\
Scheduler          & 1Cyc\tnote{g} & 1Cyc        & 1Cyc          & 1Cyc          & 1Cyc            & 1Cyc          & Cos           & WCos\tnote{g} & WCos          & WCos             \\
Batch              & 8             & 8            & 32            & 8             & 8               & 28            & 8             & 4             & 2             & 4                \\
WD                 & 1e-4          & 1e-4         & 1e-4          & 1e-4          & 1e-4            & 1e-4          & 1e-4          & 1e-3          & 6.7e-4        & 8.1e-5           \\
\addlinespace[0.3em]
\midrule
\addlinespace[0.3em]
\multicolumn{11}{l}{\textit{Regularization \& PEFT}} \\
\addlinespace[0.2em]
Dropout            & 0.5           & 0.5          & 0.1           & 0.1           & 0.1             & ---           & ---           & ---           & ---           & 0.0              \\
Grad. Clip         & ---           & ---          & ---           & ---           & ---             & ---           & ---           & 2.0           & 1.0           & 1.0              \\
PEFT               & ---           & ---          & ---           & ---           & ---             & ---           & ---           & NT\tnote{g}   & NT            & NT               \\
Adapter Dim.       & ---           & ---          & ---           & ---           & ---             & ---           & ---           & 512           & 512           & ---              \\
\bottomrule
\end{tabularx}
\begin{tablenotes}[para,flushleft]
    \item[a] DLV3+: DeepLabV3+;
    \item[b] SF: SegFormer;
    \item[c] U-R34: UNet-ResNet34;
    \item[d] LW-U: Lightweight UNet;
    \item[e] M-U: MedSAM+UNet.
    \item[h] MS: MedSAM.
    \item[f] RN: ResNet; IN: ImageNet.
    \item[g] 1Cyc: OneCycle; WCos: Warmup Cosine; Cos: Cosine; NT: Norm Tuning; WD: Weight Decay; LR: Learning Rate.
\end{tablenotes}
\end{threeparttable}
\end{table}

\subsection{Transfer Learning and Weight Initialization Strategy}
\label{sec:transfer_learning}

Fair model comparison requires consistent initialization strategies. Table \ref{tab:transfer_learning} summarizes the pretraining source, weight initialization, and fine-tuning approach for each architecture.

\begin{table}[h!]
\caption{Transfer Learning and Weight Initialization Strategy. All decoders are randomly initialized. Params indicates total model parameters.}
\label{tab:transfer_learning}
\centering
\scriptsize
\begin{tabularx}{\textwidth}{@{} l >{\centering\arraybackslash}p{1.6cm} >{\centering\arraybackslash}p{1.4cm} >{\centering\arraybackslash}p{1.4cm} >{\centering\arraybackslash}p{1.2cm} >{\centering\arraybackslash}X @{}}
\toprule
\textbf{Model} & \textbf{Pretraining} & \textbf{Encoder Frozen?} & \textbf{PEFT} & \textbf{Params (M)} & \textbf{Trainable Scope} \\
\midrule
\addlinespace[0.3em]
\multicolumn{6}{l}{\textit{\textbf{Classical Architectures}}} \\
\addlinespace[0.2em]
\midrule
FCN & ImageNet (VGG-16) & No & --- & 134.4 & Full network \\
SegNet & None & N/A & --- & 29.5 & Full network \\
\midrule
\addlinespace[0.3em]
\multicolumn{6}{l}{\textit{\textbf{CNN-based Architectures}}} \\
\addlinespace[0.2em]
\midrule
UNet & None & N/A & --- & 31.4 & Full network \\
LW-UNet & None & N/A & --- & 7.8 & Full network \\
UNet-ResNet34 & ImageNet & No & --- & 24.4 & Encoder + decoder \\
DeepLabV3+ & ImageNet & No & --- & 26.7 & Encoder + decoder + ASPP \\
\midrule
\addlinespace[0.3em]
\multicolumn{6}{l}{\textit{\textbf{Transformer-based Architecture}}} \\
\addlinespace[0.2em]
\midrule
SegFormer & ImageNet & No & --- & 3.7 & MiT encoder + MLP decoder \\
\midrule
\addlinespace[0.3em]
\multicolumn{6}{l}{\textit{\textbf{Foundation Models}}} \\
\addlinespace[0.2em]
\midrule
SAM & SA-1B & Yes & Norm Tuning & 97.3 & Norm layers + decoder + auto-prompt \\
MedSAM & SA-1B + Medical & Yes & Norm Tuning & 95.8 & Norm layers + mask decoder \\
MedSAM+UNet & SA-1B + Medical & Yes & Norm Tuning & 95.7 & Norm layers + UNet decoder \\
\bottomrule
\end{tabularx}
\end{table}

\subsubsection{CNN and Transformer Models: Full Fine-tuning}
For the encoder-based CNN architectures (\unetresnet{}, \deeplab{}) and \segformer{}, we employed full fine-tuning with ImageNet-pretrained encoders via the \texttt{segmentation-models-pytorch} library. The encoder weights were initialized with ImageNet weights but were not frozen during training, allowing the encoder to adapt its learned representations to the histopathological domain. The decoders were randomly initialized. The remaining architectures (\unet{}, \lightunet{}, \fcn{}, \segnet{}) were trained with random initialization of all weights.

\subsubsection{Foundation Models: Parameter-Efficient Fine-tuning}
For the foundation models (SAM, MedSAM, MedSAM+UNet), we adopted a parameter-efficient fine-tuning (PEFT) strategy. The image encoders were initialized from official SAM checkpoints and were frozen by default to reduce computational requirements and prevent overfitting. All three foundation models use Norm Tuning as their PEFT method, which enables gradient computation only for the normalization layers within the frozen encoder while training the mask decoder. For SAM, this was identified as the optimal method via Bayesian search, combined with a learned auto-prompting module. For MedSAM, Norm Tuning allows the normalization layers and mask decoder to adapt to the histopathological domain. For MedSAM+UNet, the SAM encoder normalization layers are tuned alongside the custom UNet-style decoder.

\subsection{Evaluation Metrics and Cross-Validation}
Our primary evaluation metric is the macro-averaged Dice Similarity Coefficient (DSC) over the three tissue classes, chosen for its robustness to class imbalance \cite{yeung2023calibrating}. We also report Intersection over Union (IoU) \cite{muller2022towards}. To assess ranking stability, we employ two cross-validation strategies: Leave-One-Out (LOOCV), a low-bias, high-variance estimator, and 3-Fold CV, a higher-bias, lower-variance estimator \cite{bhagat2025comprehensive}.

For a given ground truth segmentation mask $\mathbf{Y} \in \{0, 1, 2, 3\}^{H \times W}$ and its corresponding predicted segmentation $\mathbf{\hat{Y}} \in \{0, 1, 2, 3\}^{H \times W}$, with classes defined as $\{0: \text{Background}, 1: \text{Lumen}, 2: \text{Neointima}, 3: \text{Media}\}$, we define the fundamental confusion matrix elements for each class $c$. The indicator function $\mathbf{1}[\cdot]$ evaluates to 1 if the condition is true and 0 otherwise.

\begin{align}
\text{TP}_c &= \sum_{i,j} \mathbf{1}[\mathbf{Y}(i,j) = c \wedge \mathbf{\hat{Y}}(i,j) = c] \\
\text{FP}_c &= \sum_{i,j} \mathbf{1}[\mathbf{Y}(i,j) \neq c \wedge \mathbf{\hat{Y}}(i,j) = c] \\
\text{FN}_c &= \sum_{i,j} \mathbf{1}[\mathbf{Y}(i,j) = c \wedge \mathbf{\hat{Y}}(i,j) \neq c]
\end{align}
where $\text{TP}_c$ (True Positives) counts pixels correctly classified as class $c$, $\text{FP}_c$ (False Positives) counts pixels incorrectly predicted as class $c$, and $\text{FN}_c$ (False Negatives) counts pixels of class $c$ misclassified as other classes.

The Dice Similarity Coefficient ($DSC_c$) quantifies the spatial overlap between the predicted and ground truth segmentations for a given class $c$:
$$DSC_c = \frac{2 \cdot TP_c}{2 \cdot TP_c + FP_c + FN_c}$$
To ensure our optimization targets clinically relevant tissues, we compute a macro-averaged Dice coefficient ($DSC_{\text{macro}}$) by excluding the background class ($c=0$):
$$DSC_{\text{macro}} = \frac{1}{3} \sum_{c=1}^{3} DSC_c$$

Intersection over Union ($IoU_c$), also known as the Jaccard index, provides a complementary measure of segmentation quality for class $c$:
$$IoU_c = \frac{TP_c}{TP_c + FP_c + FN_c}$$
The relationship between DSC and IoU is direct:
$$DSC_c = \frac{2 \cdot IoU_c}{1 + IoU_c}$$

Our loss function, identified as optimal through our hyperparameter search, is a composite of Focal and Dice loss designed to balance pixel-level accuracy and region-level overlap \cite{yeung2022unified}.

\subsubsection{Focal Loss}\label{sec:focal_loss}
To mitigate the impact of severe class imbalance, we employ Focal Loss ($\mathcal{L}_{\text{focal}}$). This loss function down-weights easy examples and focuses training on hard, misclassified examples:
$$\mathcal{L}_{\text{focal}} = -\alpha (1 - p_t)^\gamma \log(p_t)$$
Here, $p_t$ represents the model's estimated probability for the ground truth class. We set $\alpha = 0.8$ for class balancing and $\gamma = 2.0$ to control the focusing strength on hard examples.

\subsubsection{Dice Loss}\label{sec:dice_loss}
The Dice Loss ($\mathcal{L}_{\text{dice}}$) directly optimizes our primary evaluation metric, promoting better overlap between predicted and ground truth masks. Its differentiable form is given by:
\begin{equation}
\mathcal{L}_{\text{dice}} = 1 - \frac{1}{C} \sum_{c=1}^{C} \frac{2 \sum_{i,j} \mathbf{Y}_c(i,j) \cdot \mathbf{\hat{Y}}_c(i,j) + \epsilon}{\sum_{i,j} \mathbf{Y}_c(i,j) + \sum_{i,j} \mathbf{\hat{Y}}_c(i,j) + \epsilon}
\end{equation}
In this formulation, $\mathbf{\hat{Y}}_c$ denotes the predicted probabilities for class $c$, $\mathbf{Y}_c$ is the one-hot encoded ground truth for class $c$, and $\epsilon = 10^{-6}$ is a small constant added to prevent division by zero. $C$ is the total number of classes.

\subsubsection{Composite Loss Function}\label{sec:composite_loss}
Our primary and most effective loss function, the Composite Loss Function ($\mathcal{L}_{\text{focal\_dice}}$), strategically combines Focal Loss and Dice Loss with adaptive weighting:
$$\mathcal{L}_{\text{focal\_dice}} = \lambda_f \mathcal{L}_{\text{focal}} + \lambda_d \mathcal{L}_{\text{dice}}$$
Here, $\lambda_f = 0.5$ and $\lambda_d = 0.5$ provide a balanced optimization of both pixel-level accuracy and region-level overlap, leading to superior segmentation results.

\subsection{Multi-Modal Explainable AI (XAI) Framework}\label{sec:xai_methods}
\begin{figure}[h!]
\centering
\includegraphics[width=1\textwidth]{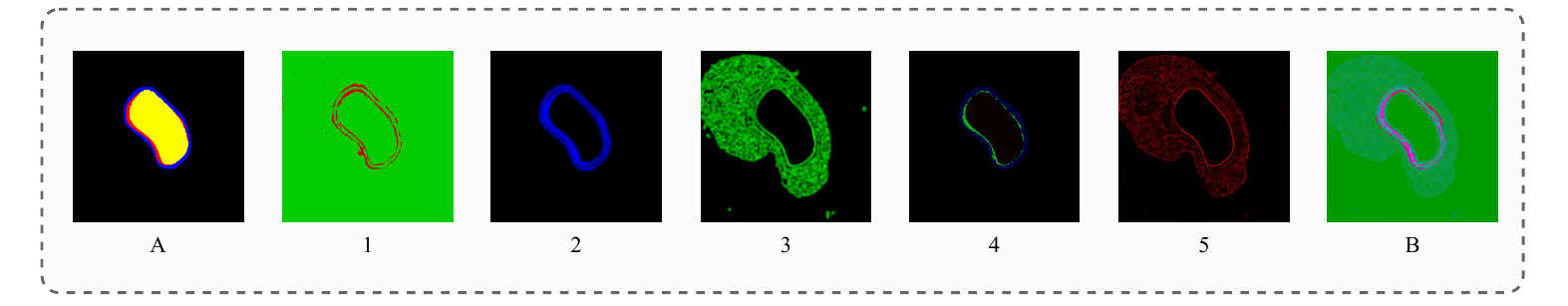}
\caption{The five-layer XAI framework. (A) A model's prediction is dissected into (1) Error Analysis, (2) Uncertainty, (3) Morphology, (4) Attention, and (5) Saliency. (B) These are synthesized into an integrated explanation, providing a multi-faceted view of the model's behavior.}
\label{fig:dataset_example_xai}
\end{figure}

Interpretability supports clinical adoption of automated segmentation. We present a framework that integrates five XAI modalities (Figure \ref{fig:dataset_example_xai}) to provide a multi-faceted diagnostic view. We apply this framework not only to explain individual predictions but also to diagnose sources of statistical instability across different data splits. An explainability framework is only useful if its outputs map directly to clinical needs. We designed our five layers to address distinct, practical questions that arise during pathological review, from assessing diagnostic accuracy to building confidence in the model's predictions. Table \ref{tab:xai_clinical_validation} provides a systematic validation of how each component serves critical clinical functions.

Mathematically, our framework generates a composite explanation map $\mathbf{E}$ from an input image $\mathbf{I}$, its ground truth $\mathbf{Y}$, and the model's prediction $\mathbf{\hat{Y}}$. This composite map is constructed as a weighted sum of five individual, distinct explanation layers $\mathbf{E}_i$:
$$\mathbf{E} = \sum_{i=1}^{5} \alpha_i \mathbf{E}_i$$
The coefficients $\alpha_i$ are adaptive and can be adjusted to emphasize different explanatory aspects based on the specific clinical question or the user's focus. The individual layers are detailed below:

\subsubsection{Layer 1: Error Analysis}\label{sec:error_analysis}
This foundational layer provides direct visual feedback on the correctness of the model's prediction. It generates a binary error map $\mathbf{E}_1$ that immediately highlights regions of agreement and disagreement between the predicted mask $\mathbf{\hat{Y}}$ and the ground truth $\mathbf{Y}$.
$$
\mathbf{E}_1(i,j) = \begin{cases}
\textcolor{green}{\blacksquare} \text{ (Green, RGB: 0, 204, 0)} & \text{if } \mathbf{\hat{Y}}(i,j) = \mathbf{Y}(i,j) \\
\textcolor{red}{\blacksquare} \text{ (Red, RGB: 204, 0, 0)} & \text{otherwise}
\end{cases}
$$
Here, \textcolor{green}{green} represents correctly classified pixels where the prediction matches the ground truth, while \textcolor{red}{red} highlights erroneous pixels where the model's prediction diverges from the expert annotation.

\subsubsection{Layer 2: Uncertainty Estimation}\label{sec:uncertainty_estimation}
This layer quantifies the model's confidence in its pixel-wise predictions. Uncertainty is often highest at critical tissue boundaries, indicating areas where the model is less certain about its classification. We compute an uncertainty map $\mathbf{E}_2$ using the pixel-wise entropy $\mathcal{H}$ of the model's class probabilities $\mathbf{P}$.
$$\mathbf{E}_2(i,j) = \mathcal{H}(\mathbf{P}(i,j)) = -\sum_{c} P_c(i,j) \log_2 P_c(i,j)$$

\subsubsection{Layer 3: Morphological Analysis}\label{sec:morphological_analysis}
To ground the explanation in clinically relevant anatomical and structural features, this layer $\mathbf{E}_3$ captures tissue-specific morphological characteristics by integrating various image processing techniques, such as texture descriptors and structural filters.

\subsubsection{Layer 4: Class-wise Attention}\label{sec:class_attention}
This layer $\mathbf{E}_4$ adaptively highlights regions corresponding to specific classes where the model exhibits suboptimal performance. The contribution of each class mask $\mathbf{M}_c$ to the overall attention map is weighted by its performance deficit, calculated as $(1 - DSC_c)$. This approach focuses the explanation on the most problematic tissue types.
$$\mathbf{E}_4(i,j) = \sum_{c=1}^{3} (1 - DSC_c) \cdot \mathbf{M}_c(i,j)$$

\subsubsection{Layer 5: Gradient-based Saliency}\label{sec:gradient_saliency}
The final layer $\mathbf{E}_5$ emphasizes the fine-grained structural edges that are often critical for accurate histological interpretation. This is achieved by computing the magnitude of the image gradient, highlighting areas of rapid intensity change.
$$\mathbf{E}_5(i,j) = \sqrt{\left(\frac{\partial \mathbf{I}}{\partial i}\right)^2 + \left(\frac{\partial \mathbf{I}}{\partial j}\right)^2}$$
This multi-modal framework provides clinicians with a detailed view of model behavior, supporting informed diagnostic decisions.

\begin{table}[h!]
\centering
\caption{Clinical Validation Framework for XAI Components. Column headers (1-5, B) refer to the components detailed in Figure \ref{fig:dataset_example_xai}: (1) Error, (2) Uncertainty, (3) Morphology, (4) Attention, (5) Saliency, and (B) Integrated.}
\label{tab:xai_clinical_validation}
\renewcommand{\arraystretch}{1.2}
\begin{tabular*}{\textwidth}{@{}l@{\extracolsep{\fill}}cccccc@{}}
\toprule
\textbf{Clinical Need} & \textbf{(1)} & \textbf{(2)} & \textbf{(3)} & \textbf{(4)} & \textbf{(5)} & \textbf{(B)} \\
\midrule
Diagnostic Accuracy     & \yes & \yes & \yes & \yes & \yes & \yes \\
Boundary Delineation    & \no  & \yes & \yes & \no  & \yes & \yes \\
Tissue Characterization & \no  & \no  & \yes & \no  & \no  & \yes \\
Quality Assurance       & \yes & \yes & \no  & \yes & \no  & \yes \\
Educational Value       & \yes & \yes & \yes & \yes & \yes & \yes \\
Clinical Confidence     & \no  & \yes & \yes & \yes & \no  & \yes \\
\bottomrule
\end{tabular*}
\end{table}

\subsection{Statistical Analysis}\label{sec:statistical_analysis}
Given the small sample size of our dataset ($N=9$), we cannot assume a normal distribution for performance scores. Therefore, our statistical analysis relies primarily on non-parametric methods to compare model performance and assess the stability of the results. All statistical analyses were conducted using Python, leveraging the \textit{pandas} library for data manipulation and \textit{Matplotlib} with \textit{Seaborn} for visualization.

To quantify the uncertainty around mean performance scores without assuming a normal distribution, we calculated 95\% bootstrap confidence intervals \cite{efron1992bootstrap}. This was achieved by resampling the cross-validation fold scores with replacement 10,000 times for each model to generate a distribution of the mean, from which the percentile-based confidence intervals were derived. The core resampling was implemented using the \texttt{resample} function from \textit{scikit-learn}\footnote{\url{https://scikit-learn.org/}} \cite{pedregosa2011scikit}.

To formally compare the models against each other across all cross-validation folds, we employed the Friedman test \cite{friedman1937use}. This non-parametric test is highly recommended for comparing multiple classifiers over multiple datasets (or, in our case, folds) \cite{demvsar2006statistical}. We then used the Nemenyi test as a post-hoc analysis to identify which specific models have statistically different average ranks. The results are visualized using Critical Difference (CD) plots. This entire analysis was performed using functions from the \textit{SciPy}\footnote{\url{https://scipy.org/}} library \cite{virtanen2020scipy}.

Finally, to assess the practical magnitude of the performance differences between model pairs, we calculated Cohen's d \cite{cohen1988statistical}. This provides a standardized measure of the difference between two means, helping to distinguish between statistical significance and practical importance.

\FloatBarrier
\section{Results}\label{sec:results}

This section details the empirical findings of our study. We first present quantitative performance under two cross-validation protocols, then examine the stability of these results through statistical analysis.

\subsection{Quantitative Performance Across Validation Protocols}
To establish a performance baseline, all optimally-configured models were evaluated using both Leave-One-Out (LOOCV) and 3-Fold cross-validation (CV).

Under the LOOCV protocol, which provides a nearly unbiased but high-variance estimate of performance, \medsam{} emerged as the apparent leader. As shown in Table \ref{tab:loocv_comprehensive_stats}, \medsam{} achieved a macro-averaged Dice Similarity Coefficient (DSC) of 0.694, followed by \segformer{} at 0.576. With the expanded set of ten models, the classical architectures (\fcn{}, \segnet{}) performed substantially worse, achieving DSC scores below 0.15.

\begin{figure}[h!]
\centering
\includegraphics[width=\textwidth]{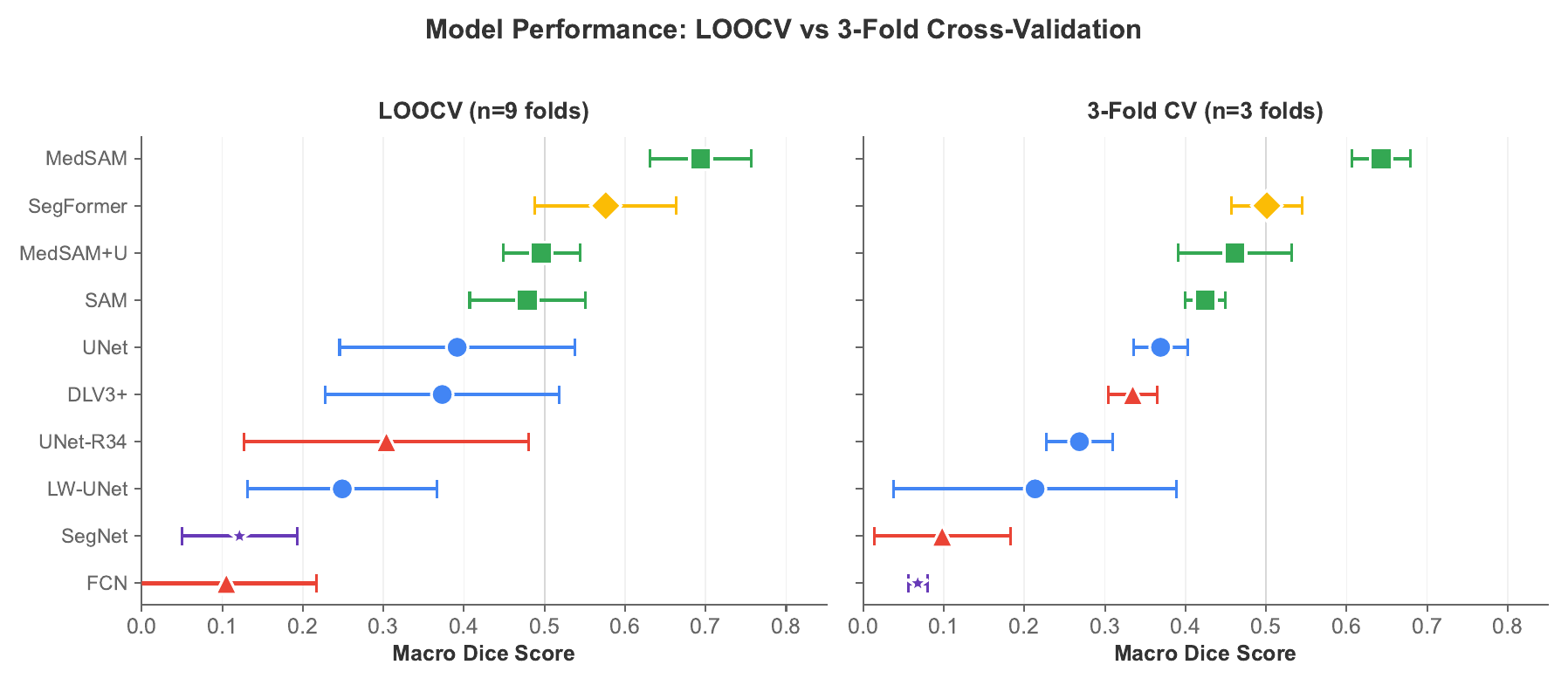}
\caption{Performance Comparison Across Validation Protocols. Side-by-side comparison of model rankings under LOOCV (left) and 3-Fold CV (right). While MedSAM leads under both protocols, substantial rank instability exists among other models---the crossing lines highlight how relative positions change between protocols.}
\label{fig:perf_comparison}
\end{figure}

Under the 3-Fold CV protocol, which has lower variance but potentially higher bias (Table \ref{tab:kfold_comprehensive_stats}), \medsam{} retained the top position (0.643 DSC), followed by \segformer{} (0.501 DSC). While \medsam{} leads in both protocols, substantial rank instability exists among other models---as visualized in Figure \ref{fig:perf_comparison}. While \medsam{} leads in both protocols, the statistical robustness of this ranking is examined in the following sections.

\begin{table}[h!]
\caption{LOOCV Performance Summary (10 Models). Results are shown as Mean $\pm$ Std Dev. Models ranked by Macro Dice score; \medsam{} achieves the highest mean.}
\label{tab:loocv_comprehensive_stats}
\centering
\setlength{\tabcolsep}{3pt}
\scriptsize
\begin{tabularx}{\textwidth}{@{} l *{2}{>{\centering\arraybackslash}X} @{}}
\toprule
\textbf{Model} & \textbf{Macro Dice $\uparrow$} & \textbf{Macro IoU $\uparrow$} \\
\midrule
\addlinespace[0.2em]
\multicolumn{3}{l}{\textit{Foundation Models \& Transformers}} \\
\addlinespace[0.1em]
\midrule
\textbf{MedSAM} & \textbf{0.694 $\pm$ 0.063} & \textbf{0.592 $\pm$ 0.058} \\
SegFormer & 0.576 $\pm$ 0.088 & 0.474 $\pm$ 0.074 \\
SAM & 0.496 $\pm$ 0.047 & 0.412 $\pm$ 0.035 \\
MedSAM+UNet & 0.479 $\pm$ 0.072 & 0.402 $\pm$ 0.068 \\
\midrule
\addlinespace[0.2em]
\multicolumn{3}{l}{\textit{Modern CNNs}} \\
\addlinespace[0.1em]
\midrule
UNet & 0.392 $\pm$ 0.146 & 0.319 $\pm$ 0.122 \\
UNet-ResNet34 & 0.373 $\pm$ 0.145 & 0.297 $\pm$ 0.132 \\
DeepLabV3+ & 0.304 $\pm$ 0.177 & 0.240 $\pm$ 0.150 \\
Lightweight UNet & 0.249 $\pm$ 0.118 & 0.200 $\pm$ 0.102 \\
\midrule
\addlinespace[0.2em]
\multicolumn{3}{l}{\textit{Classical Architectures}} \\
\addlinespace[0.1em]
\midrule
FCN & 0.121 $\pm$ 0.071 & 0.075 $\pm$ 0.055 \\
SegNet & 0.105 $\pm$ 0.111 & 0.076 $\pm$ 0.088 \\
\bottomrule
\end{tabularx}
\end{table}

\begin{table}[h!]
\caption{3-Fold Cross-Validation Performance Summary (10 Models). Results are shown as Mean $\pm$ Std Dev across 3 folds. Note the ranking changes compared to LOOCV.}
\label{tab:kfold_comprehensive_stats}
\centering
\setlength{\tabcolsep}{3pt}
\scriptsize
\begin{tabularx}{\textwidth}{@{} l *{2}{>{\centering\arraybackslash}X} @{}}
\toprule
\textbf{Model} & \textbf{Macro Dice $\uparrow$} & \textbf{Macro IoU $\uparrow$} \\
\midrule
\addlinespace[0.2em]
\multicolumn{3}{l}{\textit{Foundation Models \& Transformers}} \\
\addlinespace[0.1em]
\midrule
\textbf{MedSAM} & \textbf{0.643 $\pm$ 0.036} & \textbf{0.542 $\pm$ 0.032} \\
SegFormer & 0.501 $\pm$ 0.044 & 0.415 $\pm$ 0.040 \\
MedSAM+UNet & 0.461 $\pm$ 0.070 & 0.388 $\pm$ 0.072 \\
SAM & 0.425 $\pm$ 0.025 & 0.351 $\pm$ 0.029 \\
\midrule
\addlinespace[0.2em]
\multicolumn{3}{l}{\textit{Modern CNNs}} \\
\addlinespace[0.1em]
\midrule
UNet & 0.369 $\pm$ 0.033 & 0.289 $\pm$ 0.034 \\
DeepLabV3+ & 0.335 $\pm$ 0.030 & 0.268 $\pm$ 0.033 \\
UNet-ResNet34 & 0.268 $\pm$ 0.041 & 0.201 $\pm$ 0.041 \\
Lightweight UNet & 0.213 $\pm$ 0.176 & 0.169 $\pm$ 0.149 \\
\midrule
\addlinespace[0.2em]
\multicolumn{3}{l}{\textit{Classical Architectures}} \\
\addlinespace[0.1em]
\midrule
SegNet & 0.098 $\pm$ 0.085 & 0.069 $\pm$ 0.069 \\
FCN & 0.068 $\pm$ 0.012 & 0.036 $\pm$ 0.006 \\
\bottomrule
\end{tabularx}
\end{table}

\subsection{Analysis of Benchmark Instability}
Further analysis indicates that no single model can be reliably identified as superior, as rankings depend on the specifics of the evaluation protocol. We identified three primary sources of this instability.

\subsubsection{Sensitivity of Model Rankings to Evaluation Protocol}
The most direct evidence of instability is that the top-ranked model changes depending on the evaluation context. Table \ref{tab:illusion_definitive} summarizes the best-performing model for each metric across three contexts: the 80/20 train-test split using optimized hyperparameters, the mean LOOCV result, and the mean 3-Fold CV result. While \medsam{} wins across all evaluation protocols in this table, it is notable that during hyperparameter optimization, \unet{} achieved the highest validation Dice score (0.630) compared to \medsam{} (0.569)---yet this ranking reverses on held-out test data. This exemplifies how validation-based model selection can be misleading.

Furthermore, the substantial range between highest and lowest scores across all 10 models (e.g., 0.007--0.781 for LOOCV Dice, representing a 0.774 range) highlights the substantial performance variability within each protocol.

\begin{table}[p]
\centering
\caption{Analysis of Performance Instability Across Evaluation Contexts (10 Models). The Range ($\Delta$) column shows performance spread across all 10 models, quantifying instability magnitude.}
\label{tab:illusion_definitive}
\definecolor{rangehigh}{RGB}{234,67,53} 
\definecolor{rangemid}{RGB}{247,144,136} 
\definecolor{rangelow}{RGB}{252,201,197} 
\definecolor{rangevlow}{RGB}{253,232,230} 
\footnotesize
\begin{tabularx}{\textwidth}{@{} l l >{\raggedleft\arraybackslash}X >{\raggedleft\arraybackslash}X >{\raggedleft\arraybackslash}X @{}}
\toprule
\textbf{Metric} & \textbf{Winner} & \textbf{Highest} & \textbf{Lowest} & \textbf{Range ($\Delta$)} \\
\midrule
\addlinespace[0.2em]
\multicolumn{5}{@{}l}{\textbf{Best Single Run (80/20 Split with Optimized Hyperparameters)}} \\
\addlinespace[0.2em]
\multicolumn{5}{@{}l}{\textit{Macro-Averaged Metrics (Overall Performance)}} \\
Dice (Macro) & \medsam & 0.609 & 0.057 & 0.552 \\
IoU (Macro) & \medsam & 0.513 & 0.030 & 0.483 \\
\addlinespace[0.2em]
\multicolumn{5}{@{}l}{\textit{Class-Specific Metrics (Tissue-Level Performance)}} \\
Dice (Lumen) & \medsam & 0.941 & 0.125 & 0.816 \\
Dice (Neointima) & \medsam & 0.224 & 0.000 & 0.224 \\
Dice (Media) & \medsam & 0.662 & 0.040 & 0.622 \\
IoU (Lumen) & \medsam & 0.903 & 0.067 & 0.836 \\
IoU (Neointima) & \medsam & 0.129 & 0.000 & 0.129 \\
IoU (Media) & \medsam & 0.507 & 0.020 & 0.487 \\
\midrule
\addlinespace[0.2em]
\multicolumn{5}{@{}l}{\textbf{Leave-One-Out CV (LOOCV)}} \\
\addlinespace[0.2em]
\multicolumn{5}{@{}l}{\textit{Macro-Averaged Metrics (Overall Performance)}} \\
Dice (Macro) & \medsam & 0.781 & 0.007 & 0.774 \\
IoU (Macro) & \medsam & 0.676 & 0.003 & 0.673 \\
\addlinespace[0.2em]
\multicolumn{5}{@{}l}{\textit{Class-Specific Metrics (Tissue-Level Performance)}} \\
Dice (Lumen) & \medsam & 0.974 & 0.002 & 0.972 \\
Dice (Neointima) & \medsam & 0.715 & 0.000 & 0.715 \\
Dice (Media) & \medsam & 0.788 & 0.000 & 0.788 \\
IoU (Lumen) & \medsam & 0.950 & 0.001 & 0.949 \\
IoU (Neointima) & \medsam & 0.582 & 0.000 & 0.582 \\
IoU (Media) & \medsam & 0.656 & 0.000 & 0.656 \\
\midrule
\addlinespace[0.2em]
\multicolumn{5}{@{}l}{\textbf{3-Fold CV (K-Fold)}} \\
\addlinespace[0.2em]
\multicolumn{5}{@{}l}{\textit{Macro-Averaged Metrics (Overall Performance)}} \\
Dice (Macro) & \medsam & 0.688 & 0.022 & 0.667 \\
IoU (Macro) & \medsam & 0.584 & 0.011 & 0.573 \\
\addlinespace[0.2em]
\multicolumn{5}{@{}l}{\textit{Class-Specific Metrics (Tissue-Level Performance)}} \\
Dice (Lumen) & \medsam & 0.958 & 0.001 & 0.957 \\
Dice (Neointima) & \medsam & 0.451 & 0.002 & 0.449 \\
Dice (Media) & \medsam & 0.712 & 0.054 & 0.658 \\
IoU (Lumen) & \medsam & 0.921 & 0.000 & 0.920 \\
IoU (Neointima) & \medsam & 0.325 & 0.001 & 0.323 \\
IoU (Media) & \medsam & 0.570 & 0.028 & 0.542 \\
\bottomrule
\end{tabularx}
\end{table}

\subsubsection{Mean Scores Obscure High Fold-to-Fold Variance}
\begin{figure}[h!]
\centering
\includegraphics[width=\linewidth]{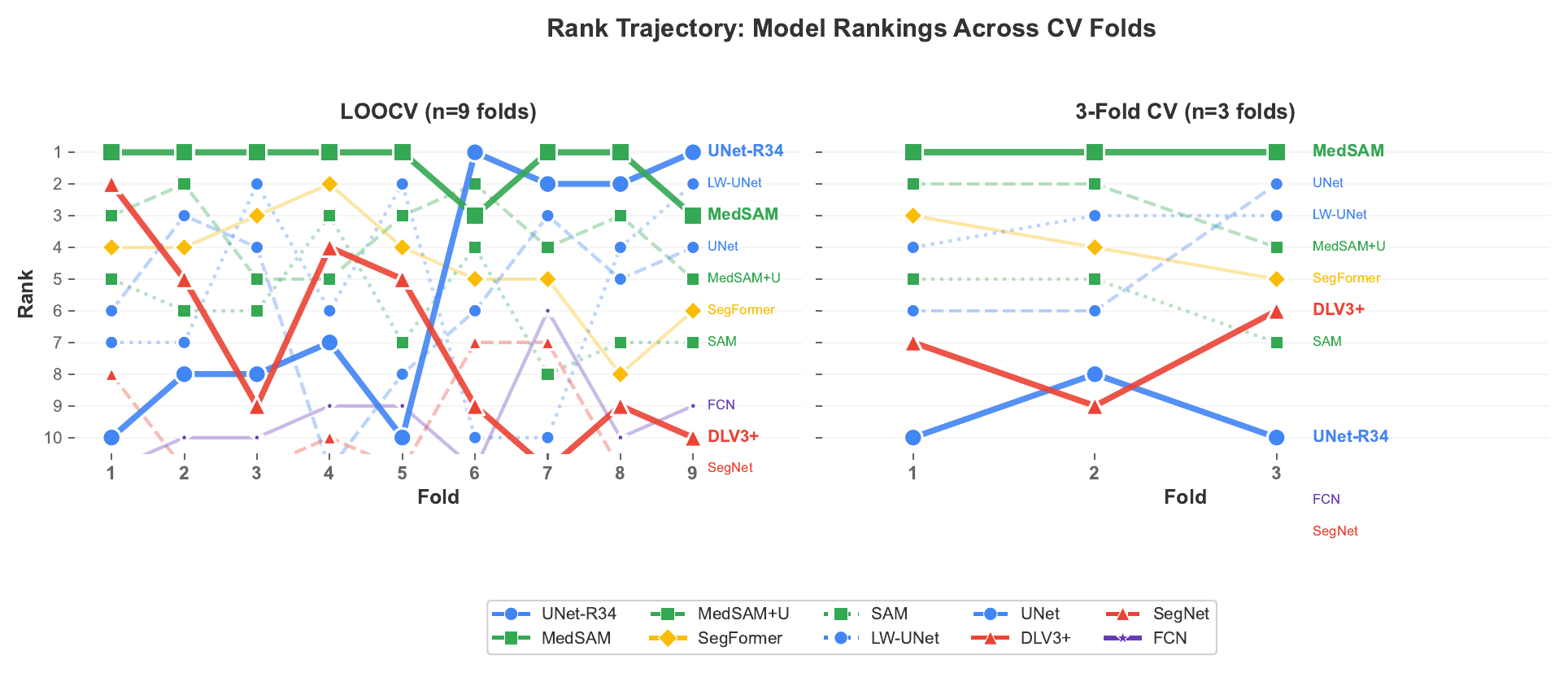}
\caption{Model Rank Stability in 3-Fold vs. Leave-one-out cross-validation (LOOCV). The plots track the performance rank of each model (1=Best) across the validation folds for 3-Fold CV (left) and LOOCV (right). The rank trajectories reveal substantial fold-to-fold volatility, particularly under LOOCV.}
\label{fig:loocv_ranking_traj_supp}
\end{figure}

Average performance metrics, while useful, can mask substantial variation across individual data splits. This is visualized in Figure \ref{fig:loocv_ranking_traj_supp}, which plots the performance rank of each model on each fold of the cross-validation. The left panel, showing 3-Fold CV, exhibits relatively consistent rankings. By contrast, the right panel for LOOCV reveals substantial rank variability. For example, while \medsam{} maintains a stable 1st-place rank across all folds, models such as \unetresnet{} and \deeplab{} exhibit rank swings of up to 7 positions between folds. This indicates that a high average score does not guarantee reliable performance on any single, unseen sample, an important concern for clinical applications.

\subsubsection{Statistical Analysis of Performance Differences}
\begin{figure}[h!]
\centering
\includegraphics[width=\linewidth]{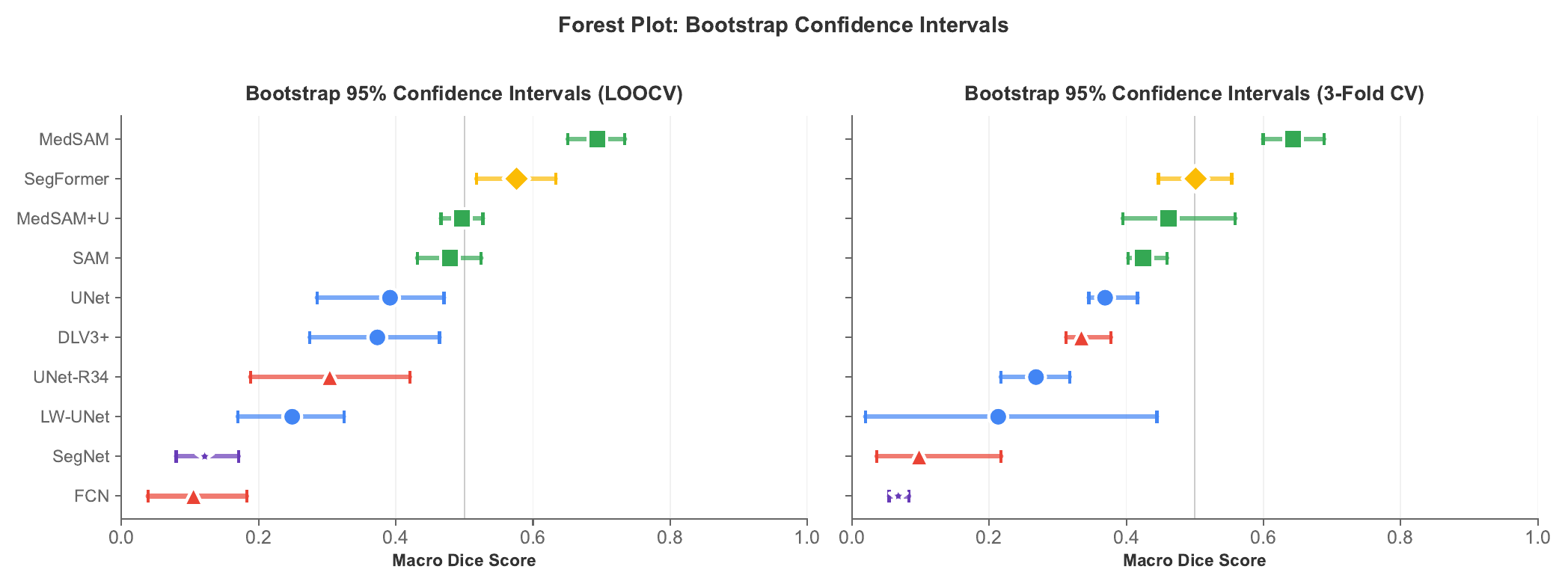}
\caption{Bootstrap 95\% Confidence Intervals for Macro-Dice Score. The plots show the mean Dice score (marker) and the bootstrap 95\% confidence interval (10,000 resamples) for each model under 3-Fold CV (left) and LOOCV (right). Under LOOCV, MedSAM's CI is non-overlapping with all other models, while substantial overlap among 2nd--5th ranked models indicates their fine-grained ranking is not statistically meaningful.}
\label{fig:bootstrap_ci_forest}
\end{figure}

To rigorously test the stability of the observed performance, we conducted a multi-faceted statistical analysis. First, to move beyond simple mean scores, we estimated the uncertainty of each model's performance using bootstrap resampling. By creating 10,000 new sets of fold scores via sampling with replacement, we constructed 95\% confidence intervals (CIs) for the true mean macro-Dice score. The results, shown in Table \ref{tab:supp_bootstrap} and visualized in Figure \ref{fig:bootstrap_ci_forest}, reveal a pattern of uncertainty. Under the LOOCV protocol, MedSAM's CI ([0.651, 0.734]) is non-overlapping with all other models, providing statistical evidence that it is the best-performing architecture. However, among models ranked 2nd through 5th, a chain of pairwise CI overlaps prevents confident fine-grained ranking: SegFormer ([0.517, 0.633]) overlaps with SAM ([0.466, 0.527]) and MedSAM+UNet ([0.431, 0.525]), both of which in turn overlap with UNet ([0.285, 0.470]). This pattern of overlapping confidence intervals indicates that, while MedSAM is statistically separable as the top performer, the ranking among the remaining competitive models is not statistically robust.

\begin{table}[h!]
\centering
\caption{Bootstrap 95\% Confidence Intervals for Macro-Dice Score (10 Models)}
\label{tab:supp_bootstrap}
\sisetup{table-format=1.3}
\scriptsize
\setlength{\tabcolsep}{6pt}
\begin{tabular}{@{} l S c S c @{}}
\toprule
& \multicolumn{2}{c}{\textbf{LOOCV}} & \multicolumn{2}{c}{\textbf{3-Fold CV}} \\
\cmidrule(lr){2-3} \cmidrule(lr){4-5}
\textbf{Model} & {\textbf{Mean}} & {\textbf{95\% CI}} & {\textbf{Mean}} & {\textbf{95\% CI}} \\
\midrule
\textbf{MedSAM} & \textbf{0.694} & {[0.651, 0.734]} & \textbf{0.643} & {[0.600, 0.688]} \\
SegFormer & 0.576 & {[0.517, 0.633]} & 0.501 & {[0.447, 0.554]} \\
SAM & 0.496 & {[0.466, 0.527]} & 0.425 & {[0.403, 0.459]} \\
MedSAM+UNet & 0.479 & {[0.431, 0.525]} & 0.461 & {[0.395, 0.559]} \\
UNet & 0.392 & {[0.285, 0.470]} & 0.369 & {[0.345, 0.416]} \\
UNet-ResNet34 & 0.373 & {[0.274, 0.464]} & 0.268 & {[0.217, 0.317]} \\
DeepLabV3+ & 0.304 & {[0.188, 0.420]} & 0.335 & {[0.312, 0.378]} \\
Lightweight UNet & 0.249 & {[0.170, 0.324]} & 0.213 & {[0.019, 0.445]} \\
FCN & 0.121 & {[0.080, 0.171]} & 0.068 & {[0.054, 0.083]} \\
SegNet & 0.105 & {[0.038, 0.183]} & 0.098 & {[0.036, 0.218]} \\
\bottomrule
\end{tabular}
\end{table}

\begin{figure}[h!]
\centering
\includegraphics[width=\linewidth]{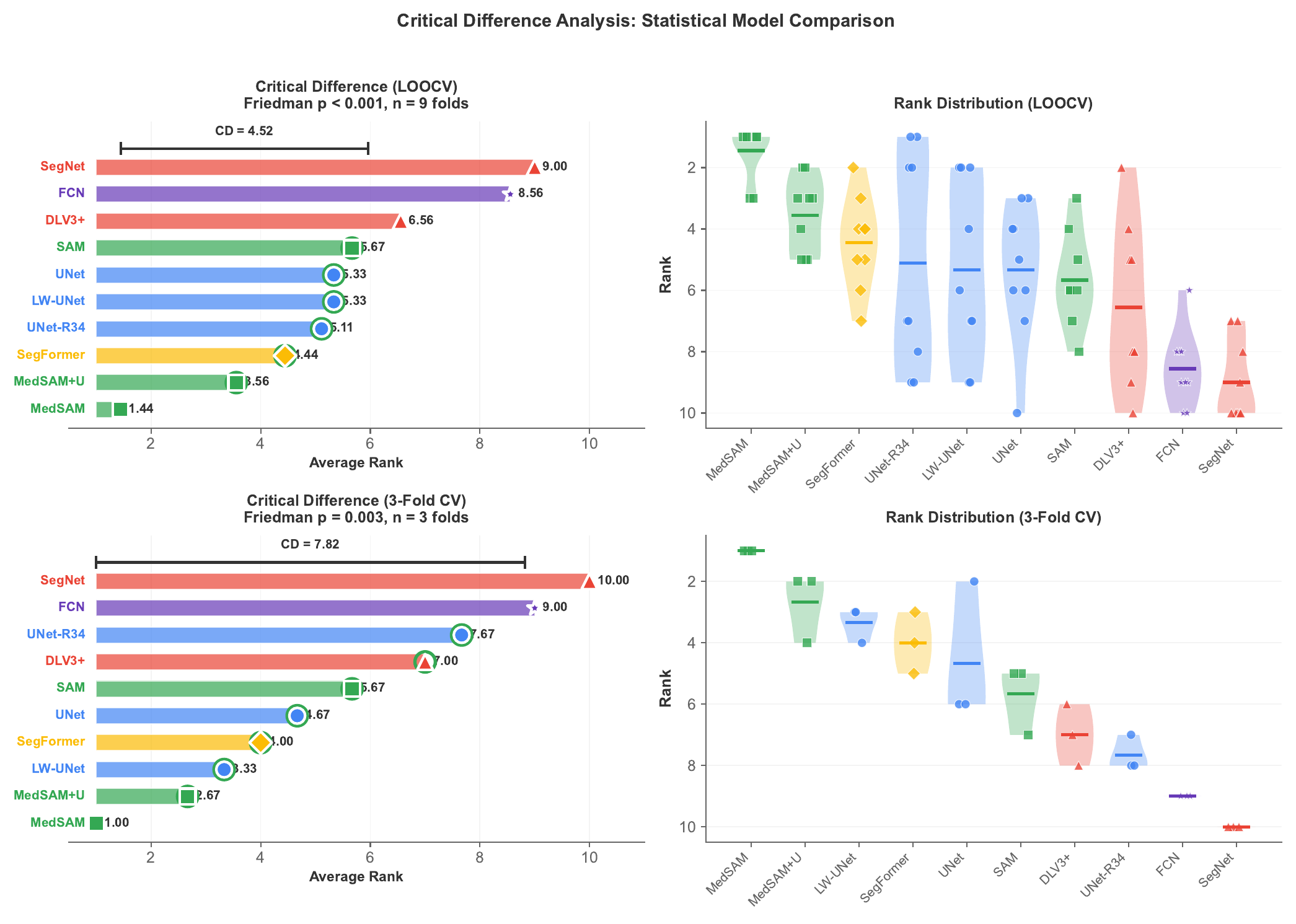}
\caption{Critical Difference Analysis of Model Ranks. This analysis compares average model ranks under LOOCV (top) and 3-Fold CV (bottom). The plots on the left are Critical Difference (CD) diagrams from the Friedman and Nemenyi tests; models connected by a solid bar are not statistically different. The plots on the right show the distribution of ranks for each model across the validation folds. Both protocols yield significant Friedman tests (LOOCV: $p < 0.001$; 3-Fold CV: $p=0.003$), confirming overall rank differences. However, the large critical difference in the 3-Fold analysis (CD $= 7.82$) means the Nemenyi post-hoc test finds no statistically significant pairwise difference among the top-performing models.}
\label{fig:critical_analysis}
\end{figure}

Next, we performed a non-parametric rank comparison using the Friedman test and Nemenyi post-hoc test, visualized with Critical Difference plots in Figure \ref{fig:critical_analysis}. This analysis directly compares the models' rankings across all folds. Both the LOOCV ($p < 0.001$) and 3-Fold CV ($p=0.003$) analyses yield statistically significant Friedman tests, confirming that some rank differences exist across the full set of ten models. However, the large critical difference threshold in the 3-Fold analysis (CD $= 7.82$) means that the Nemenyi post-hoc test connects all top-tier models, providing strong evidence that they are statistically indistinguishable from one another despite the overall test significance. We note that with only three observations per model, the 3-Fold analysis has limited statistical power for distinguishing ten models; the LOOCV analysis---with nine observations per model---provides the more reliable basis for pairwise comparison.

\begin{figure}[h!]
    \centering
    \includegraphics[width=1\linewidth]{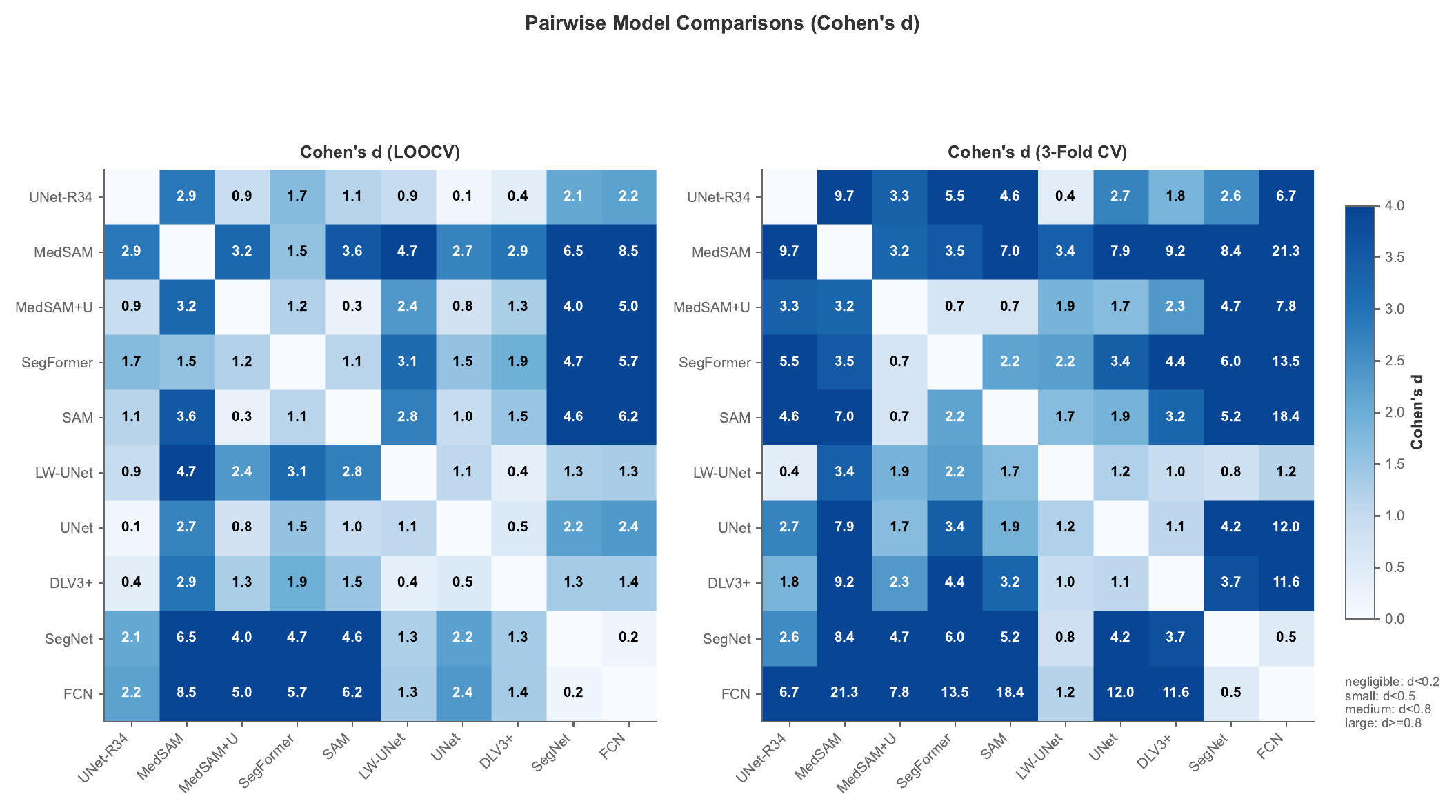}
    \caption{Pairwise Effect Size (Cohen's d) Matrix for all 10 models. The heatmap visualizes the practical significance of performance differences between model pairs. Large effect sizes ($d > 0.8$) dominate comparisons between architectural families (e.g., foundation models vs.\ classical architectures), while within-family comparisons yield smaller effects, indicating that fine-grained rankings within a family are not practically meaningful.}
    \label{fig:cohen_d_effect_size}
\end{figure}

Finally, to distinguish between statistical significance and practical importance, we calculated the effect size (Cohen's d) for all pairwise comparisons (Table \ref{tab:supp_cohens_d} and Figure \ref{fig:cohen_d_effect_size}). This analysis reveals a structured pattern: large effect sizes ($d > 0.8$) dominate comparisons between architectural families (e.g., foundation models vs.\ classical CNNs), while within-family comparisons (e.g., among CNN variants or among foundation models) yield smaller effects. The rank volatility, detailed in Table~\ref{tab:supp_rank_instability}, confirms that no consistent, reproducible model hierarchy exists, with high rank standard deviations across folds (Table~\ref{tab:supp_rank_instability}). These results indicate that while differences between architectural families are meaningful, fine-grained rankings within a family are not statistically robust.

\begin{table}[h!]
\begin{threeparttable}
\caption{Pairwise Effect Size (Cohen's d) for LOOCV Macro-Dice Scores (10 Models). Positive values indicate the row model outperforms the column model.}
\label{tab:supp_cohens_d}
\centering
\scriptsize
\setlength{\tabcolsep}{4pt}
\begin{tabular*}{\textwidth}{@{\extracolsep{\fill}} l c c c c c c c c c @{}}
\toprule
\textbf{vs.} & \textbf{SF} & \textbf{SAM} & \textbf{M-U}\tnote{b} & \textbf{UNet} & \textbf{U-R34}\tnote{c} & \textbf{DLV3+}\tnote{a} & \textbf{LW-U}\tnote{d} & \textbf{FCN} & \textbf{SegNet} \\
\midrule
\textbf{MedSAM} & 1.55 & 3.56 & 3.19 & 2.69 & 2.87 & 2.94 & 4.72 & 8.54 & 6.52 \\
\textbf{SegFormer} & -- & 1.13 & 1.21 & 1.53 & 1.69 & 1.95 & 3.15 & 5.69 & 4.70 \\
\textbf{SAM} & -- & -- & 0.29 & 0.97 & 1.14 & 1.49 & 2.76 & 6.20 & 4.57 \\
\textbf{M-UNet}\tnote{b} & -- & -- & -- & 0.76 & 0.92 & 1.30 & 2.36 & 5.00 & 3.99 \\
\textbf{UNet} & -- & -- & -- & -- & 0.13 & 0.54 & 1.08 & 2.35 & 2.21 \\
\textbf{U-R34}\tnote{c} & -- & -- & -- & -- & -- & 0.43 & 0.94 & 2.20 & 2.07 \\
\textbf{DLV3+}\tnote{a} & -- & -- & -- & -- & -- & -- & 0.36 & 1.35 & 1.34 \\
\textbf{LW-U}\tnote{d} & -- & -- & -- & -- & -- & -- & -- & 1.31 & 1.26 \\
\textbf{FCN} & -- & -- & -- & -- & -- & -- & -- & -- & 0.17 \\
\bottomrule
\end{tabular*}
\begin{tablenotes}[para,flushleft]
\footnotesize
\item[a] DLV3+: DeepLabV3+ \quad
\item[b] M-UNet: MedSAM+UNet \quad
\item[c] U-R34: UNet-ResNet34 \quad
\item[d] LW-U: Lightweight UNet
\end{tablenotes}
\end{threeparttable}
\end{table}

\begin{table}[h!]
\centering
\begin{threeparttable}
\caption{Comparative Rank Instability Analysis Across Cross-Validation Protocols (10 Models)}
\label{tab:supp_rank_instability}
\sisetup{round-mode=places, round-precision=2}
\scriptsize
\setlength{\tabcolsep}{3pt}
\begin{tabular*}{\textwidth}{@{\extracolsep{\fill}}l *{5}{c} *{5}{c} @{}}
\toprule
& \multicolumn{5}{c}{\textbf{LOOCV (High Variance)}} & \multicolumn{5}{c}{\textbf{3-Fold CV (Lower Variance)}} \\
\cmidrule(lr){2-6} \cmidrule(lr){7-11}
\textbf{Model} & $\bar{R}$ & R & $\sigma$ & \rotatebox{90}{Top-1} & \rotatebox{90}{Top-2} & $\bar{R}$ & R & $\sigma$ & \rotatebox{90}{Top-1} & \rotatebox{90}{Top-2} \\
\midrule
MedSAM & 1.00 & 0.0 & 0.00 & 100.0 & 100.0 & 1.00 & 0.0 & 0.00 & 100.0 & 100.0 \\
SegFormer & 2.33 & 1.0 & 0.50 & 0.0 & 66.7 & 2.33 & 1.0 & 0.58 & 0.0 & 66.7 \\
SAM & 3.78 & 4.0 & 1.30 & 0.0 & 11.1 & 4.33 & 1.0 & 0.58 & 0.0 & 0.0 \\
MedSAM+UNet & 4.67 & 5.0 & 1.50 & 0.0 & 11.1 & 3.67 & 4.0 & 2.08 & 0.0 & 33.3 \\
UNet & 5.78 & 7.0 & 2.11 & 0.0 & 0.0 & 4.67 & 3.0 & 1.53 & 0.0 & 0.0 \\
UNet-ResNet34 & 5.78 & 7.0 & 2.44 & 0.0 & 11.1 & 7.67 & 1.0 & 0.58 & 0.0 & 0.0 \\
DeepLabV3+ & 6.56 & 6.0 & 2.24 & 0.0 & 0.0 & 6.00 & 4.0 & 2.00 & 0.0 & 0.0 \\
LW-UNet & 7.44 & 4.0 & 1.42 & 0.0 & 0.0 & 6.33 & 2.0 & 1.15 & 0.0 & 0.0 \\
FCN & 8.67 & 3.0 & 1.00 & 0.0 & 0.0 & 9.00 & 0.0 & 0.00 & 0.0 & 0.0 \\
SegNet & 9.00 & 3.0 & 1.12 & 0.0 & 0.0 & 10.00 & 0.0 & 0.00 & 0.0 & 0.0 \\
\bottomrule
\end{tabular*}
\begin{tablenotes}[para,flushleft]
\item[] \footnotesize \textit{Note:} $\bar{R}$ is the mean rank, R is the rank range (Max - Min), $\sigma$ is the rank standard deviation, and Top-1/Top-2 show the percentage of folds where the model ranked 1st or within top 2.
\end{tablenotes}
\end{threeparttable}
\end{table}

\subsection{Qualitative Consistency Despite Quantitative Variability}
\begin{figure}[p]
\centering
\includegraphics[width=\textwidth,height=0.82\textheight,keepaspectratio]{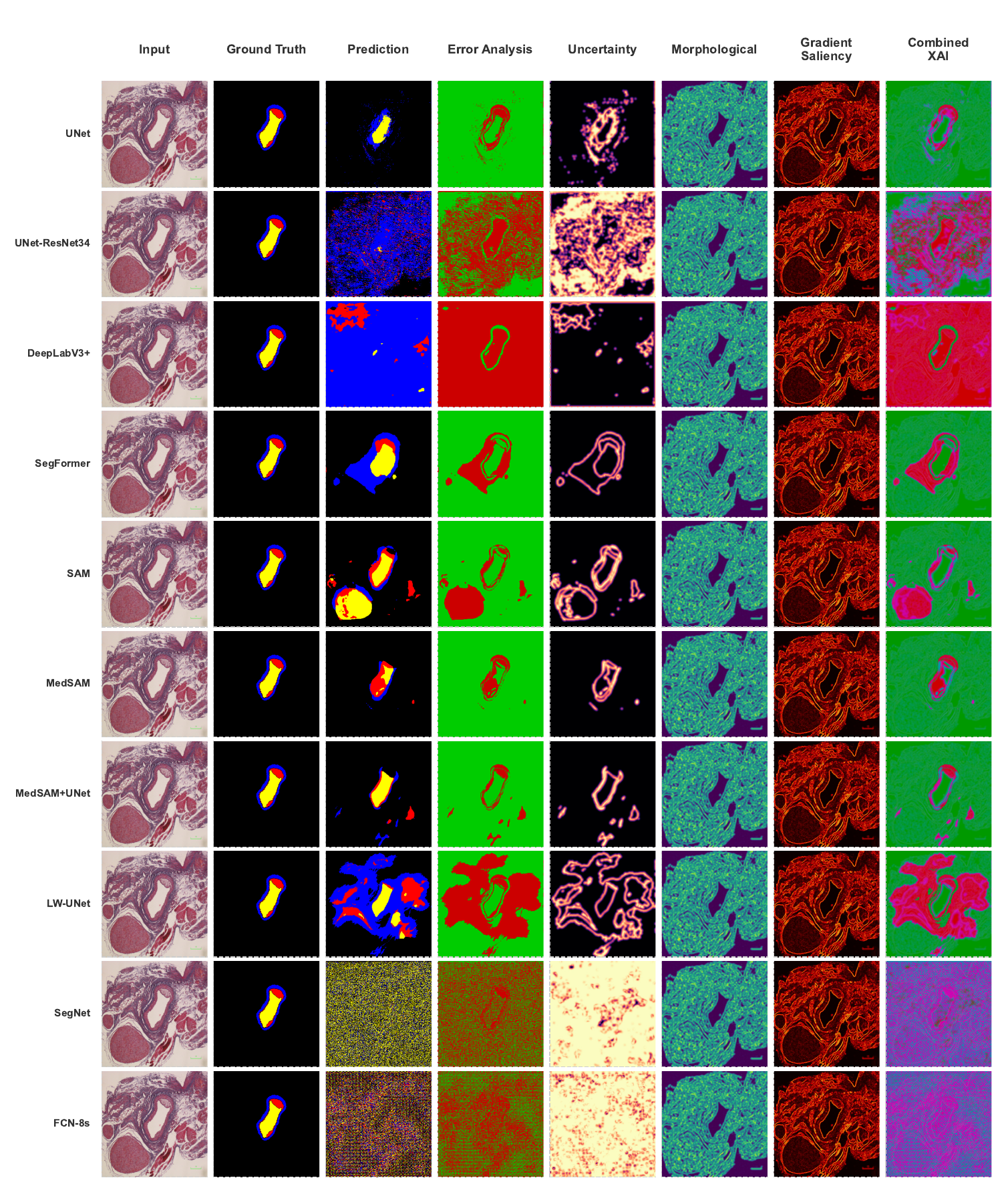}
\caption{Multi-modal XAI analysis of all ten DS1-trained architectures evaluated on a representative DS2 sample under distribution shift. Each row represents one model; columns show (left to right): input image, ground truth, model prediction, error analysis, predictive uncertainty, morphological gradient, gradient saliency, and combined XAI overlay. Models exhibit varying degrees of generalization, from accurate boundary-level predictions (UNet, SegFormer) to moderate degradation (MedSAM variants) to scattered, noisy outputs(SegNet, FCN)}
\label{fig:qualitative_xai_comparison}
\end{figure}

To complement the quantitative analysis, we conducted a multi-modal XAI analysis by applying all ten DS1-trained architectures to a representative DS2 sample under distribution shift (Figure \ref{fig:qualitative_xai_comparison}). The 10-row $\times$ 8-column grid displays, for each model: the input histological image, the ground truth mask, the model's prediction, an error analysis map, a predictive uncertainty map, a morphological gradient, a gradient-based saliency map, and a combined XAI overlay. The results reveal varying degrees of generalization across architectures. UNet and SegFormer produce predictions closest to the ground truth, with errors confined primarily to tissue boundaries. MedSAM and MedSAM+UNet maintain recognizable segmentation structure with moderate boundary errors. UNet-ResNet34 and SAM show noisier predictions than their in-distribution ranking would suggest. DeepLabV3+ exhibits substantial
overprediction of the media class, and LW-UNet displays multi-class confusion. SegNet and FCN produce scattered, noisy outputs with widespread
error and elevated uncertainty. These results demonstrate that models with comparable in-distribution performance on DS1 can exhibit substantially different generalization behavior, suggesting that out-of-distribution evaluation provides a more informative basis for model selection than in-distribution metrics alone.

\subsection{Quantitative Analysis of XAI Stability}
\begin{figure}[htbp!]
\centering
\includegraphics[width=1\textwidth]{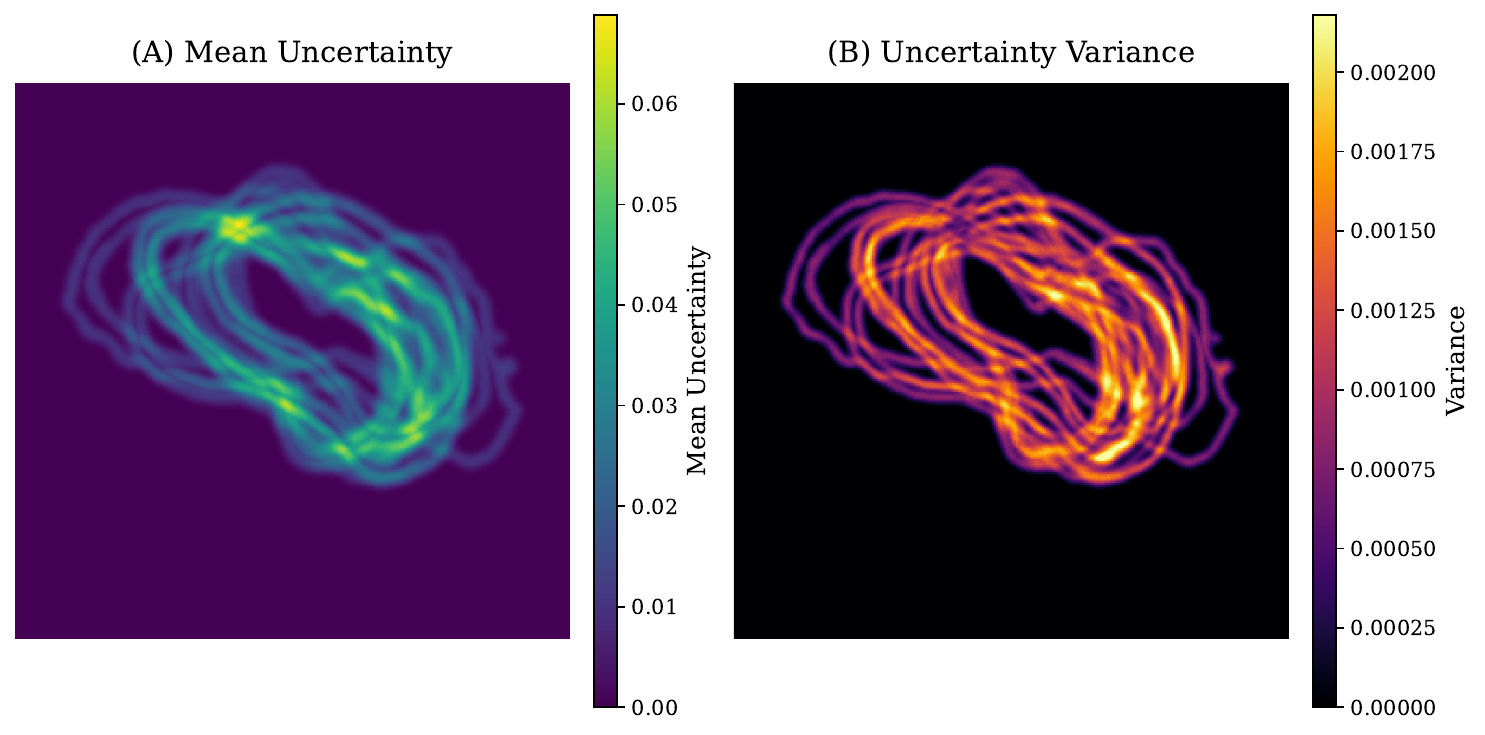}
\caption{Quantitative XAI Stability Analysis. (A) The mean uncertainty map, averaged across all 9 LOOCV folds, showing that uncertainty is consistently highest at tissue boundaries. (B) The per-pixel variance of uncertainty across the 9 folds. The extremely low variance (dark color) provides quantitative evidence that the model's confidence and reasoning were highly stable, despite fluctuating performance metrics.}
\label{fig:xai_stability}
\end{figure}

To move beyond a purely qualitative assessment of this observation, we conducted a quantitative analysis of the stability of the models' reasoning. We used the generated uncertainty maps from the 9 folds of the LOOCV protocol as a proxy for the model's confidence. For each top-performing model, we calculated the pixel-wise mean and variance of these maps. The results for the top-performing models are shown in Figure \ref{fig:xai_stability}.

The mean uncertainty map (Figure \ref{fig:xai_stability}A) confirms that, on average, the model's uncertainty is highest at the boundaries between tissue types, which is clinically expected. Figure \ref{fig:xai_stability}B displays the variance map, which is nearly black across the entire image. This indicates a quantitatively minimal variance in uncertainty across the 9 independent folds. This indicates that while the quantitative performance metrics were unstable, the model's underlying reasoning was highly stable and consistent. The instability is therefore confined to superficial metric noise at the boundaries, not a fundamental disagreement in model competence.

\FloatBarrier
\subsection{Generalization and Dataset-Dependence of Model Rankings}\label{sec:generalization}

To test whether the observed benchmark instabilities translate to real-world deployment scenarios, we conducted two complementary experiments on an independent generalization dataset (DS2) comprising $N=153$ images from different tissue preparations and magnifications than the training data (DS1). First, we evaluated all ten DS1-trained models directly on DS2 to assess robustness under distribution shift (Sections \ref{sec:dist_shift_evidence}--\ref{sec:ranking_inversions}). Second, we trained all models from scratch on DS2 at varying sample sizes ($N=9, 25, 50, 100, 150$) and tested on three held-out DS2 images, revealing dataset-specific ranking hierarchies that differ from those observed on DS1 (Section \ref{sec:ds2_indistribution}).

\begin{figure}[h!]
\centering
\includegraphics[width=\textwidth]{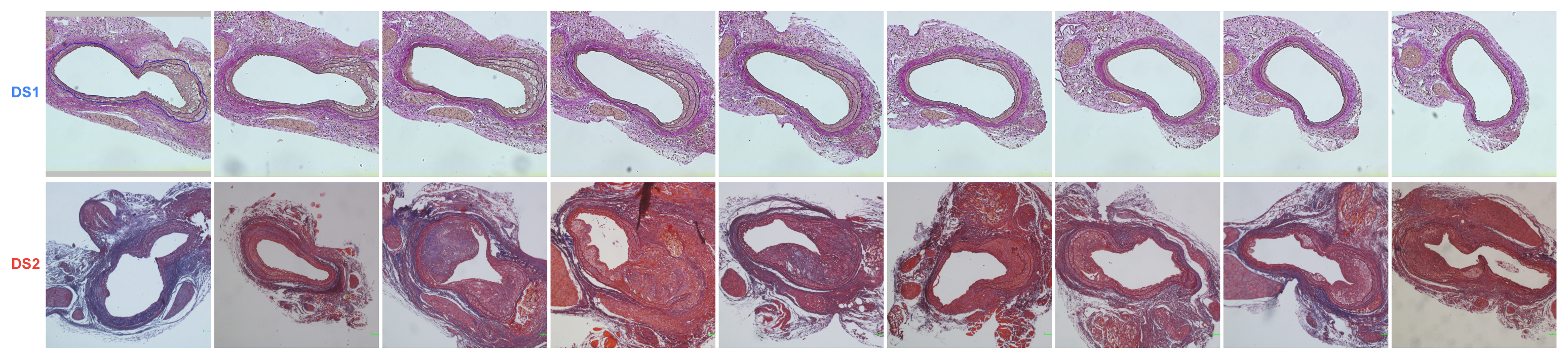}
\caption{Visual comparison of representative samples from DS1 (top row) and DS2 (bottom row). The two datasets exhibit substantial differences in staining protocols, tissue morphology, and imaging conditions, establishing DS2 as a genuine out-of-distribution evaluation scenario.}
\label{fig:dataset_comparison}
\end{figure}

Figure \ref{fig:dataset_comparison} illustrates the visual differences between the two datasets. DS1 images exhibit consistent H\&E staining with well-defined vessel morphology, whereas DS2 images display markedly different staining intensity, tissue architecture, and imaging conditions, confirming that DS2 constitutes a genuine out-of-distribution challenge rather than a simple extension of DS1.

\subsubsection{Evidence of Distribution Shift}\label{sec:dist_shift_evidence}

\begin{figure}[h!]
\centering
\includegraphics[width=0.9\textwidth]{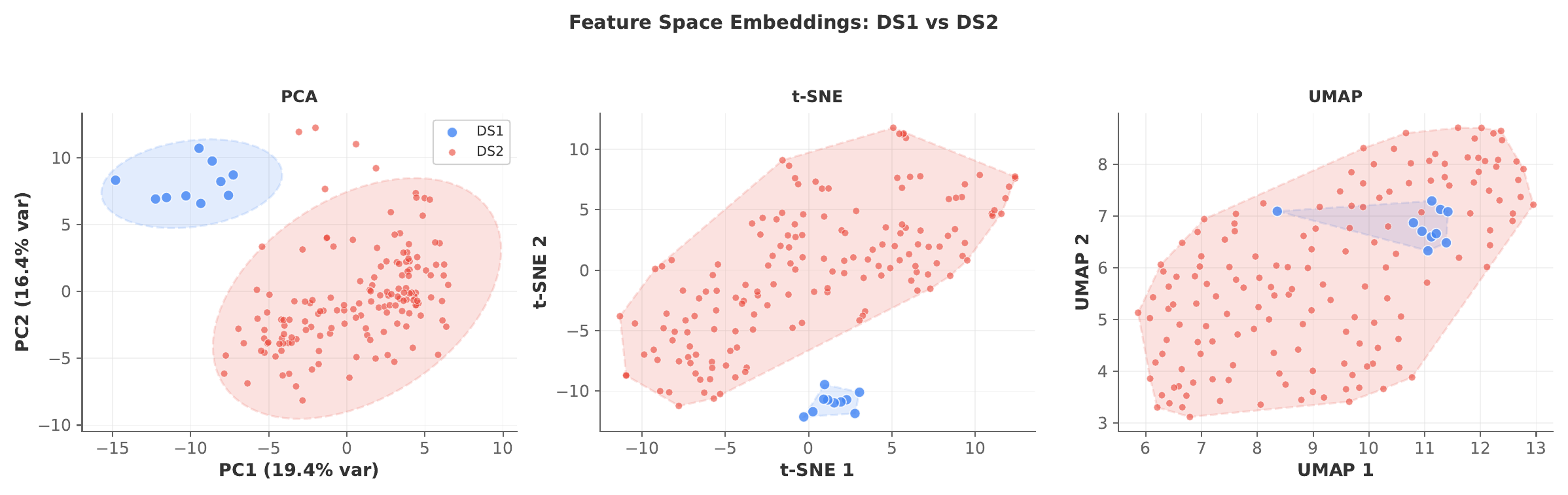}
\caption{t-SNE visualization of feature embeddings showing distribution shift between training data (DS1, $N=9$) and generalization data (DS2, $N=153$). The clear cluster separation indicates substantial domain shift in imaging conditions.}
\label{fig:distribution_shift}
\end{figure}

Figure \ref{fig:distribution_shift} visualizes the distribution shift between DS1 and DS2 using t-SNE embeddings of features extracted from a pre-trained model. The clear separation between the two datasets confirms that DS2 represents an out-of-distribution evaluation scenario.

\subsubsection{Sample Size Sensitivity Analysis (Out-of-Distribution)}

\begin{figure}[h!]
\centering
\includegraphics[width=\textwidth]{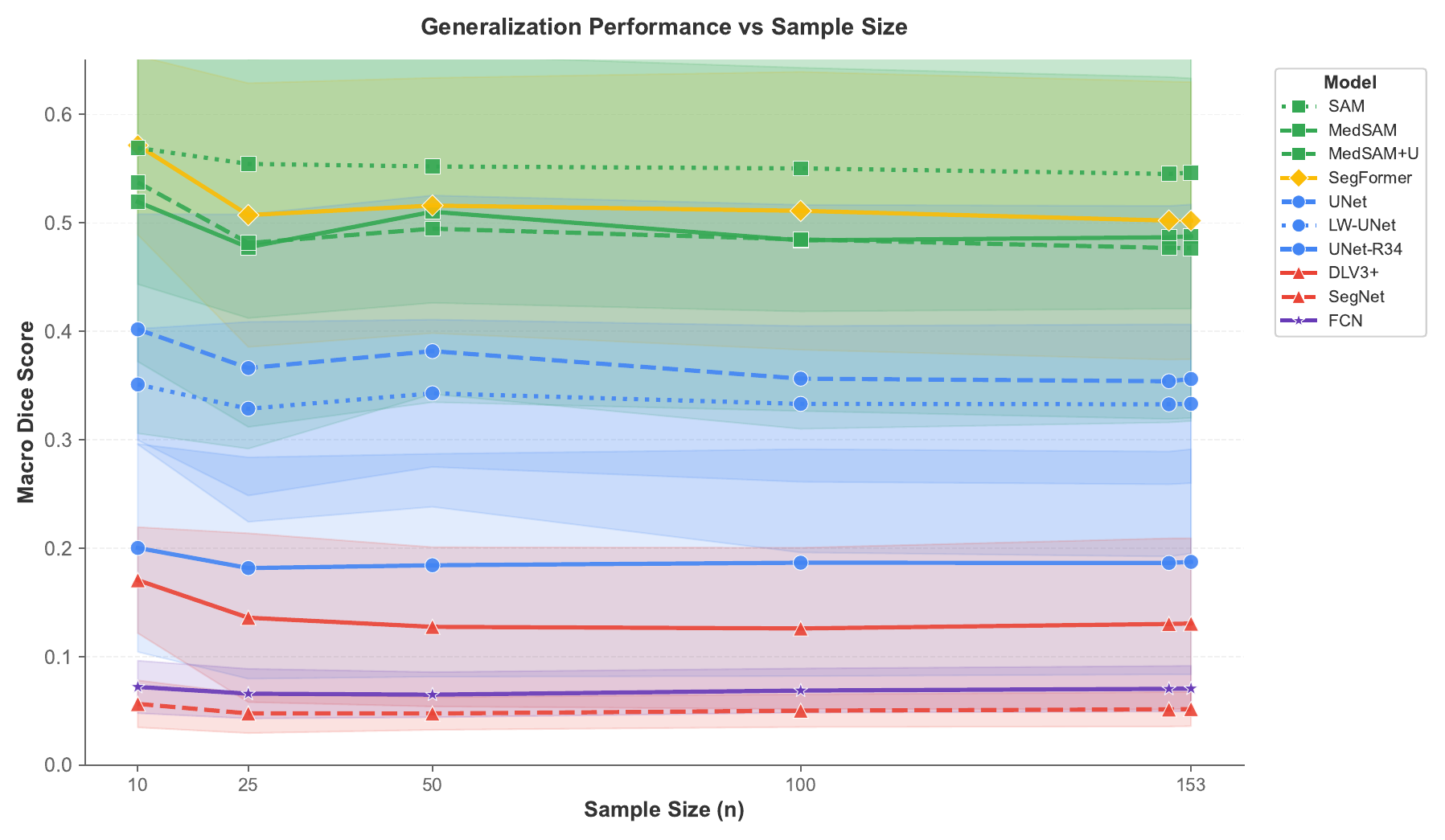}
\caption{Out-of-distribution generalization performance vs.\ evaluation sample size. DS1-trained models were evaluated on nested subsets of DS2 ($n=10$ to $153$). Foundation models maintain stable performance across sample sizes, while classical architectures show persistent failure under distribution shift.}
\label{fig:generalization_sample_size}
\end{figure}

To understand how model rankings evolve with increasing evaluation sample sizes under distribution shift, we evaluated DS1-trained models on nested subsets of DS2 ($n=10, 25, 50, 100, 153$ samples). As shown in Figure \ref{fig:generalization_sample_size}, model rankings are unstable at small sample sizes ($n \leq 25$) but stabilize as sample size increases. Foundation models (MedSAM, SAM) maintain consistent performance across sample sizes, whereas classical architectures (FCN, SegNet) fail to generalize under distribution shift.

\subsubsection{Ranking Inversions Under Distribution Shift}\label{sec:ranking_inversions}

Model rankings under distribution shift differ from in-distribution rankings. While \medsam{} maintains top performance on both DS1 and DS2, the relative ranking of other models changes considerably. For instance, \segformer{}, ranked 2nd on DS1, maintains a similar relative position on DS2. However, \unet{} variants, which showed competitive DS1 performance, suffer substantial degradation under shift. Similarly, classical architectures that showed some learning on DS1 fail almost completely on DS2.

Foundation models pre-trained on diverse distributions show better generalization, suggesting that pre-training strategy may matter more than architecture choice in data-scarce domains.

\begin{figure}[h!]
\centering
\includegraphics[width=\textwidth]{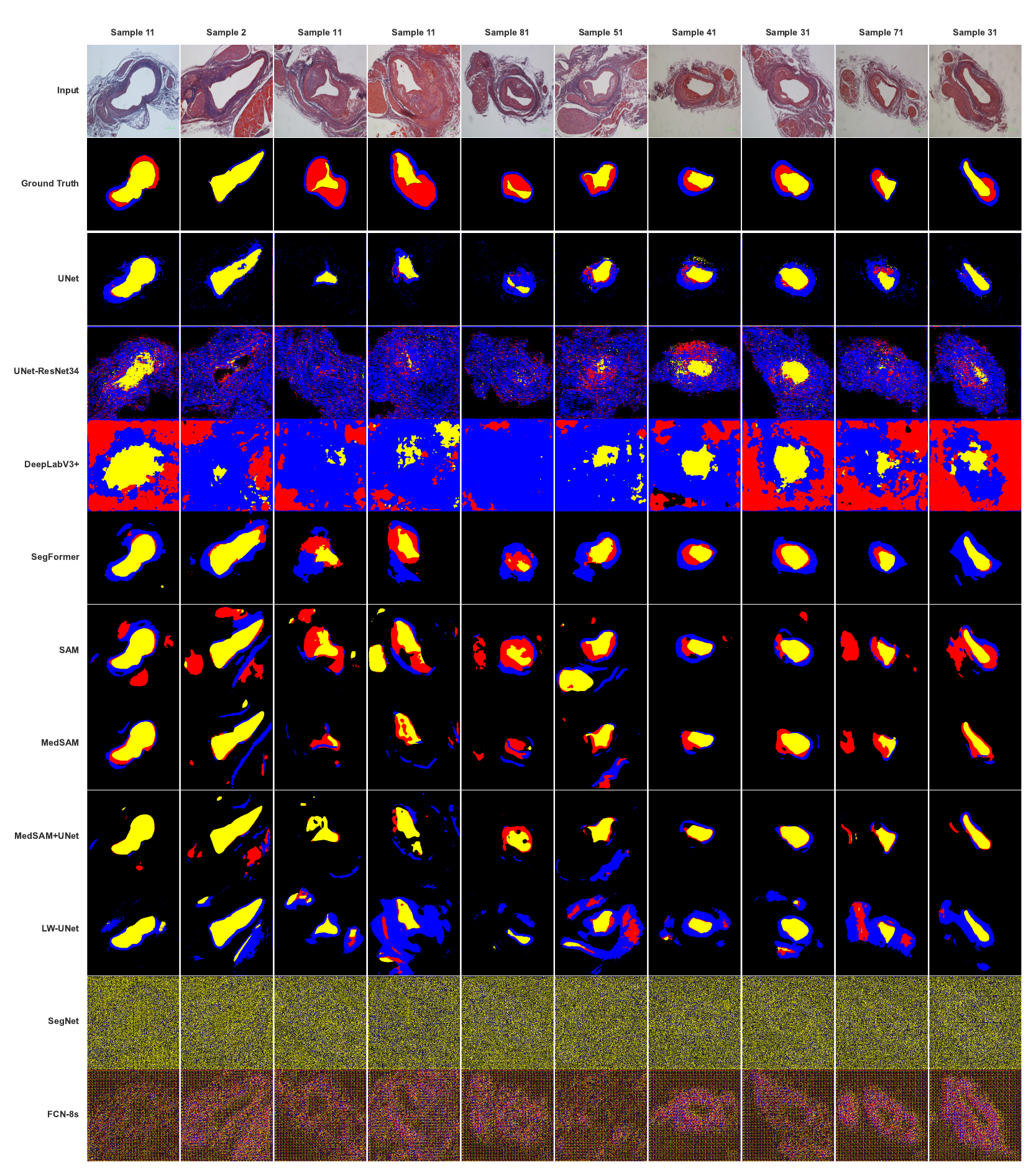}
\caption{Qualitative inference results of DS1-trained models on DS2 under distribution shift. Rows represent the input image, ground truth, and predictions from each of the ten architectures; columns span representative DS2 samples. Foundation models and top encoders maintain reasonable segmentation, while classical architectures (SegNet, FCN) produce near-uniform or random outputs, visually confirming the ranking inversions observed in quantitative analysis.}
\label{fig:inference_comparison_grid}
\end{figure}

Figure \ref{fig:inference_comparison_grid} provides visual evidence of these ranking inversions. The grid shows DS1-trained models applied to a representative sample of DS2 images spanning the full dataset. SegFormer and DeepLabV3+ maintain recognizable segmentation structure across samples, whereas SegNet and FCN produce near-uniform class predictions, confirming their complete failure under distribution shift. LW-UNet exhibits inconsistent behavior---producing reasonable segmentations on some samples but fragmented outputs on others---highlighting the unpredictability of generalization for architectures without diverse pre-training.

\subsubsection{In-Distribution Learning Curves on DS2}\label{sec:ds2_indistribution}

The preceding sections examined how DS1-trained models perform under distribution shift. A complementary question is: do the same ranking hierarchies emerge when models are trained \textit{directly} on DS2? To answer this, we trained all ten architectures on DS2 subsets of increasing size ($N=9, 25, 50, 100, 150$), using three held-out DS2 images as a fixed test set. For $N \leq 50$, we used 5-fold cross-validation; for $N \geq 100$, an 80/20 train-validation split. All models used the same hyperparameters optimized on DS1 to isolate dataset effects from tuning effects.

Table \ref{tab:ds2_indistribution} reports the mean macro Dice score across the three test images for each model and dataset size. The results show a ranking hierarchy that differs from DS1.

\begin{table}[h!]
\caption{DS2 In-Distribution Performance: Mean Macro Dice Score. Models trained on DS2 subsets ($N=9$ to $150$) and tested on three held-out DS2 images. Bold indicates best performance per column.}
\label{tab:ds2_indistribution}
\centering
\scriptsize
\begin{tabularx}{\textwidth}{@{} l *{5}{>{\centering\arraybackslash}X} @{}}
\toprule
\textbf{Model} & \textbf{N=9} & \textbf{N=25} & \textbf{N=50} & \textbf{N=100} & \textbf{N=150} \\
\midrule
\addlinespace[0.2em]
\multicolumn{6}{l}{\textit{Foundation Models \& Transformers}} \\
\addlinespace[0.1em]
\midrule
SegFormer & 0.717 & 0.657 & 0.847 & 0.896 & \textbf{0.906} \\
MedSAM & 0.779 & 0.839 & 0.846 & 0.857 & 0.870 \\
MedSAM+UNet & 0.812 & 0.833 & 0.839 & 0.859 & 0.864 \\
SAM & 0.821 & 0.792 & 0.836 & 0.859 & 0.866 \\
\midrule
\addlinespace[0.2em]
\multicolumn{6}{l}{\textit{Modern CNNs}} \\
\addlinespace[0.1em]
\midrule
UNet & 0.840 & 0.820 & 0.837 & 0.889 & 0.886 \\
UNet-ResNet34 & 0.769 & \textbf{0.862} & \textbf{0.865} & 0.876 & 0.892 \\
DeepLabV3+ & \textbf{0.843} & 0.838 & 0.852 & \textbf{0.891} & 0.896 \\
Lightweight UNet & 0.605 & 0.810 & 0.787 & 0.879 & 0.904 \\
\midrule
\addlinespace[0.2em]
\multicolumn{6}{l}{\textit{Classical Architectures}} \\
\addlinespace[0.1em]
\midrule
FCN & 0.450 & 0.549 & 0.562 & 0.726 & 0.828 \\
SegNet & 0.051 & 0.067 & 0.618 & 0.634 & 0.586 \\
\bottomrule
\end{tabularx}
\end{table}

The results demonstrate dataset-specific model rankings. At $N=9$ on DS2, \deeplab{} (0.843) and \unet{} (0.840) lead the ranking---a different hierarchy from DS1, where \medsam{} (0.694) dominated under LOOCV. Foundation models, which were the clear winners on DS1, rank only 3rd--5th on DS2 at the same sample size.

The scaling behavior of different architectures further distinguishes them. \segformer{} exhibits classic data-hungry transformer behavior: it ranks near the bottom at $N=9$ (0.717, rank 7/10) but rises to the top at $N=150$ (0.906, rank 1/10), a +26\% improvement. Similarly, \lightunet{} improves from 0.605 to 0.904 (+49\%). This contrasts with foundation models, which show only modest improvement (+5--12\% from $N=9$ to $N=150$). \sam{}, \medsam{}, and \hybrid{} converge to a narrow band around 0.86--0.87 at $N=150$, despite starting from different points. Their pre-training provides a strong initialization that helps at small $N$ but offers diminishing returns as in-distribution data grows.

At the other end of the spectrum, classical architectures perform poorly. \segnet{} achieves only 0.051 Dice at $N=9$, approaching chance-level performance---and never exceeds 0.634 even at $N=100$. \fcn{} similarly struggles (0.450 at $N=9$), confirming that these older architectures lack the inductive biases necessary for learning from minimal data. Finally, the data also independently confirm the stability threshold identified in the DS1 analysis (Section \ref{sec:ablation}): rankings are volatile at $N \leq 25$ (e.g., \segformer{} drops from rank 7 to rank 8 between $N=9$ and $N=25$) but stabilize at $N \geq 50$.

\begin{figure}[p]
\centering
\includegraphics[width=\textwidth,height=0.78\textheight,keepaspectratio]{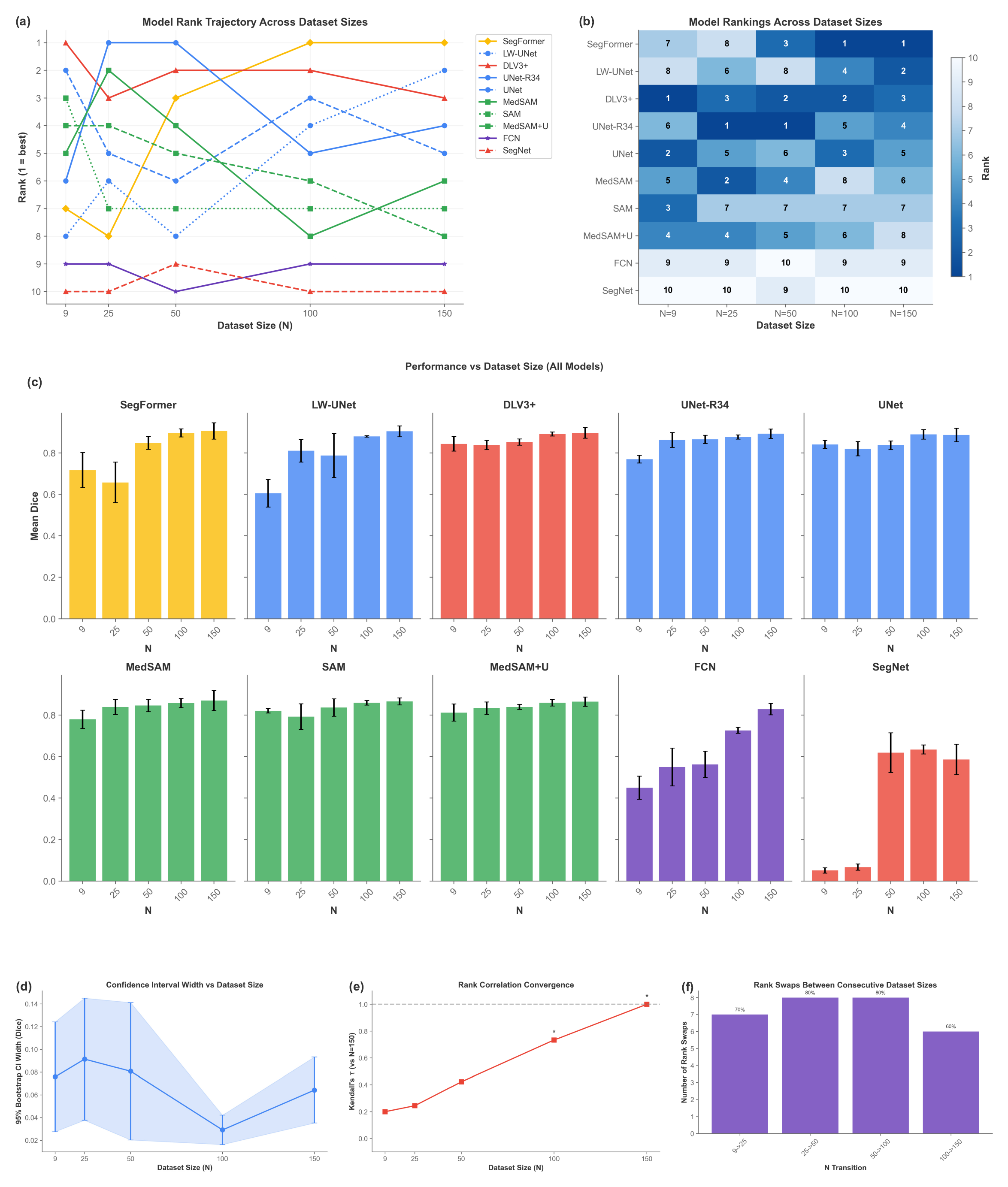}
\caption{Comprehensive DS2 in-distribution analysis across dataset sizes ($N=9$ to $150$). (a) Rank trajectory showing frequent rank crossings at small $N$ that stabilize at $N \geq 50$. (b) Rank heatmap providing a compact view of all models' rankings across dataset sizes. (c) Performance vs.\ dataset size for each architecture, revealing distinct scaling behaviors. (d) Bootstrap 95\% CI width decreasing with sample size, confirming tighter estimates at larger $N$. (e) Kendall's $\tau$ rank correlation convergence toward the final ($N=150$) ranking. (f) Number of rank swaps between consecutive dataset sizes, with a sharp decline after $N=50$ confirming the stability threshold.}
\label{fig:ds2_rank_trajectory}
\end{figure}

\begin{figure}[p]
\centering
\includegraphics[width=\textwidth,height=0.82\textheight,keepaspectratio]{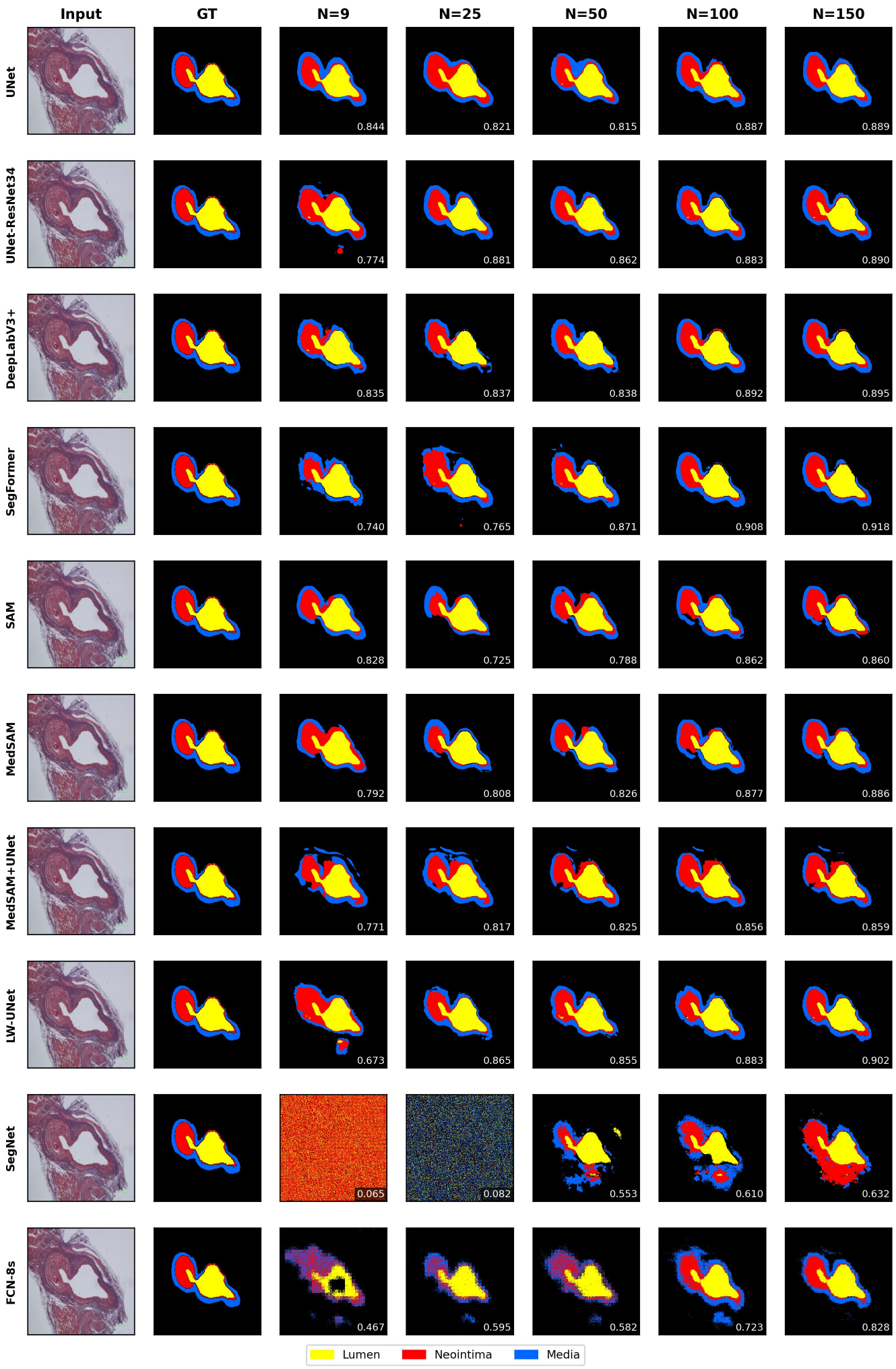}
\caption{Qualitative predictions on a held-out DS2 test image across all ten models and dataset sizes ($N=9$ to $150$). Each row shows one architecture with Dice scores annotated per panel. Columns represent increasing training data. At $N=9$, several models produce fragmented segmentations; by $N=150$, most models converge to visually accurate predictions. Classical architectures (SegNet, FCN) show persistent failure even at larger $N$, while SegFormer exhibits the largest improvement from 0.740 to 0.918.}
\label{fig:ds2_inference_grid}
\end{figure}

Figure \ref{fig:ds2_rank_trajectory} provides a six-panel analysis of these scaling dynamics. The rank trajectory (panel a) and rank heatmap (panel b) visualize the frequent crossing patterns at small $N$ that gradually converge as data increases. The performance-vs-dataset-size small multiples (panel c) reveal architecture-specific scaling curves, with SegFormer and LW-UNet showing steep improvement while foundation models plateau early. The confidence interval width (panel d) decreases monotonically with $N$, confirming that larger samples yield tighter estimates. Kendall's $\tau$ correlation with the final ranking (panel e) reaches near-perfect agreement by $N=100$, and the number of rank swaps between consecutive dataset sizes (panel f) drops sharply after $N=50$, independently confirming the stability threshold identified on DS1. The qualitative predictions in Figure \ref{fig:ds2_inference_grid} provide visual confirmation: at $N=9$, several models produce fragmented or missing segmentations, while at $N=150$ most models achieve visually accurate delineation of all tissue classes.

\subsubsection{Synthesis: Domain vs.\ Architecture}\label{sec:synthesis}

The preceding experiments reveal that model selection cannot be captured by a single leaderboard. The out-of-distribution results (Sections \ref{sec:dist_shift_evidence}--\ref{sec:ranking_inversions}) demonstrate that foundation models perform better when deployed on unseen data distributions, maintaining reasonable Dice scores where conventional architectures degrade. However, the in-distribution DS2 results (Section \ref{sec:ds2_indistribution}) show that this advantage is specifically about pre-training diversity, not inherent architectural superiority. When trained directly on DS2, conventional architectures like \deeplab{} and \segformer{} match or exceed foundation model performance.

This distinction has direct practical implications. When distribution shift is expected, foundation models offer a clear advantage through their diverse pre-training. Distribution shifts often occur in clinical deployment where training data may come from one institution and test data from another. When sufficient in-distribution data is available ($N \geq 50$), task-specific architectures like \segformer{} or \deeplab{} can achieve superior performance with far fewer computational resources. The `best'' model is therefore not a fixed property of an architecture but a function of both the available data and the expected deployment conditions.

\FloatBarrier
\subsection{Ablation Studies}\label{sec:ablation}

To systematically quantify sources of experimental variance beyond data splits, we conducted 190+ ablation experiments examining augmentation strategies, input resolution, random seed effects, and hyperparameter sensitivity.

\subsubsection{Data Augmentation and Seed Stability}

Figure \ref{fig:ablation_aug_seed} summarizes two important sources of variance. First, we tested 10 different augmentation presets (10 experiments $\times$ 10 models = 100 total runs), ranging from no augmentation to aggressive geometric and color transformations. The variance introduced by augmentation choice was substantial---for some models, the difference between best and worst augmentation preset exceeded 0.15 Dice points. Second, we trained each model with 5 different random seeds (5 seeds $\times$ 10 models = 50 runs). Seed-induced variance was smaller than augmentation variance but still meaningful, with standard deviations ranging from 0.03 to 0.15 Dice points across models, with foundation models exhibiting the least seed sensitivity (0.03--0.06) and classical/transformer architectures the most (up to 0.11 for SegNet and 0.15 for SegFormer).

\begin{figure}[h!]
\centering
\includegraphics[width=\textwidth]{assets/ablation_aug_seed_combined.pdf}
\caption{Ablation study: Data augmentation and seed stability. Left: Performance variance across 10 augmentation presets per model. Right: Performance variance across 5 random seeds per model. Both sources introduce variance comparable to architectural differences.}
\label{fig:ablation_aug_seed}
\end{figure}

Beyond demonstrating that augmentation introduces variance, the heatmap in Figure \ref{fig:ablation_aug_seed}A reveals architecture-specific augmentation preferences that yield practical guidance for practitioners.

Foundation models are augmentation-robust: \medsam{} is nearly insensitive to augmentation choice (Dice range: 0.67--0.70 across all 10 presets), achieving competitive performance even without augmentation. This robustness likely stems from the diverse pre-training distribution of the ViT encoder, which already encodes invariances that augmentation would otherwise need to teach. By contrast, CNNs benefit most from geometric augmentation. \unet{} improved from 0.41 (no augmentation) to 0.66 with strong geometric transforms (+61\%), and \deeplab{} peaked at 0.66 with the same preset. Geometric augmentations (rotations, flips, affine transforms) effectively expand the limited training distribution for architectures that lack built-in spatial invariances; for CNN-based architectures, we recommend geometric augmentation at medium-to-strong intensity.

\segformer{} uniquely prefers color augmentation. Unlike CNNs, it achieved its best performance with weak color augmentation (0.63) while strong geometric transforms degraded it to 0.34. The hierarchical transformer encoder, which already captures spatial relationships through self-attention, benefits more from staining variability simulation than from geometric perturbations; we therefore recommend color-based augmentation for transformer architectures. Finally, lightweight architectures are highly sensitive to augmentation choice: \lightunet{} exhibited substantial instability, with Dice scores ranging from 0.04 (color medium) to 0.44 (full medium)---a 10$\times$ performance ratio from augmentation choice alone. This suggests that under-parameterized models lack the capacity to simultaneously learn the task and cope with aggressive augmentation, making augmentation selection important for lightweight deployment scenarios.

These findings provide architecture-family-specific augmentation recommendations rather than treating augmentation as a universal preprocessing step.

\subsubsection{Resolution Sensitivity}

\begin{figure}[h!]
\centering
\includegraphics[width=\textwidth]{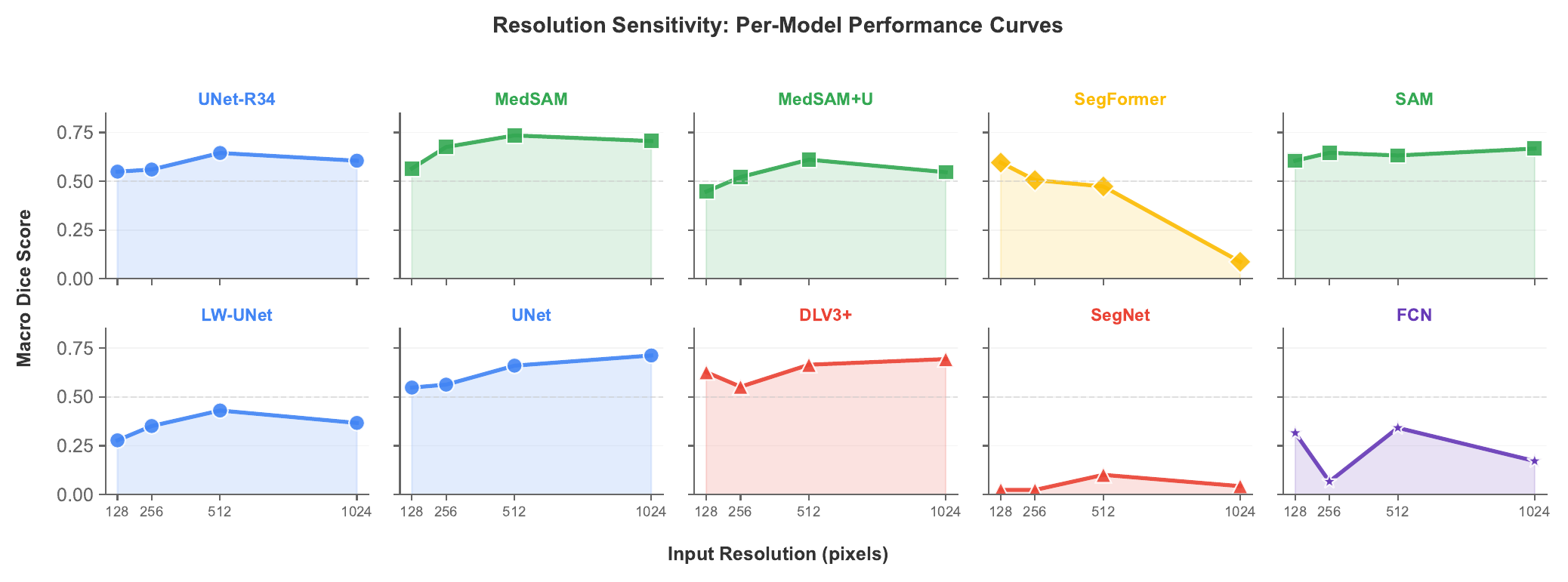}
\caption{Resolution sensitivity analysis. Performance across input resolutions (128--1024 pixels) reveals architecture-specific optimal operating points, with no universal resolution preference across model families.}
\label{fig:resolution}
\end{figure}

We evaluated all models at four input resolutions (128, 256, 512, 1024 pixels), with results shown in Figure \ref{fig:resolution}. Foundation models generally benefit from higher resolutions (512--1024px), with SAM peaking at 1024px and MedSAM at 512px. CNN architectures show mixed resolution preferences: some (UNet, DeepLabV3+) peak at 1024px, while others (UNet-ResNet34, Lightweight UNet) perform best at 512px. SegFormer uniquely peaks at 128px, suggesting its transformer architecture captures sufficient context even at low resolution. Ultra-low resolution (128px) reduces performance for most models.

\subsubsection{Hyperparameter Sensitivity}

\begin{figure}[h!]
\centering
\includegraphics[width=\textwidth]{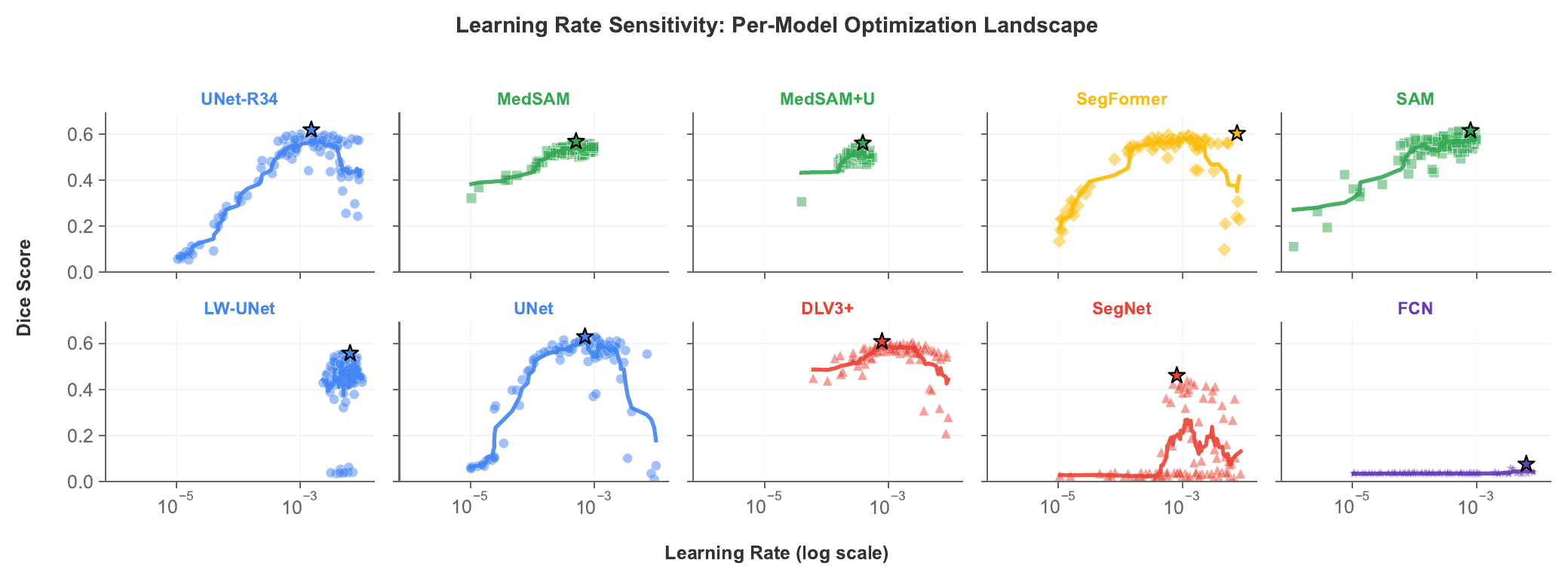}
\caption{Learning Rate Sensitivity Analysis. Performance distributions across learning rate ranges for each model, revealing architecture-specific optimal ranges. Foundation models (SAM, MedSAM) show narrow optimal ranges around $10^{-4}$, while classical architectures tolerate broader ranges.}
\label{fig:hp_learning_rate}
\end{figure}

\begin{figure}[h!]
\centering
\includegraphics[width=\textwidth]{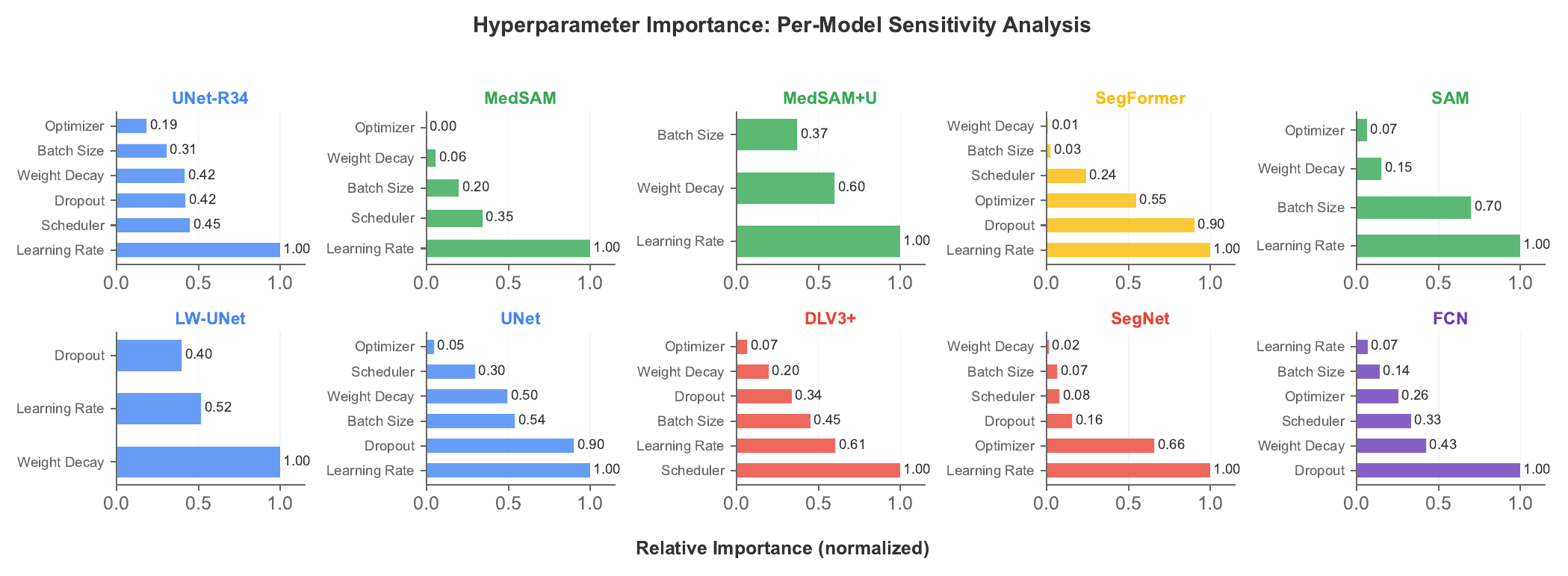}
\caption{Hyperparameter Importance Analysis. Relative importance of hyperparameters computed via Spearman correlation with validation performance. Learning rate dominates for most models, but foundation models show greater sensitivity to weight decay, while classical architectures are more sensitive to dropout and optimizer choice.}
\label{fig:hp_importance}
\end{figure}

\begin{figure}[h!]
\centering
\includegraphics[width=\textwidth]{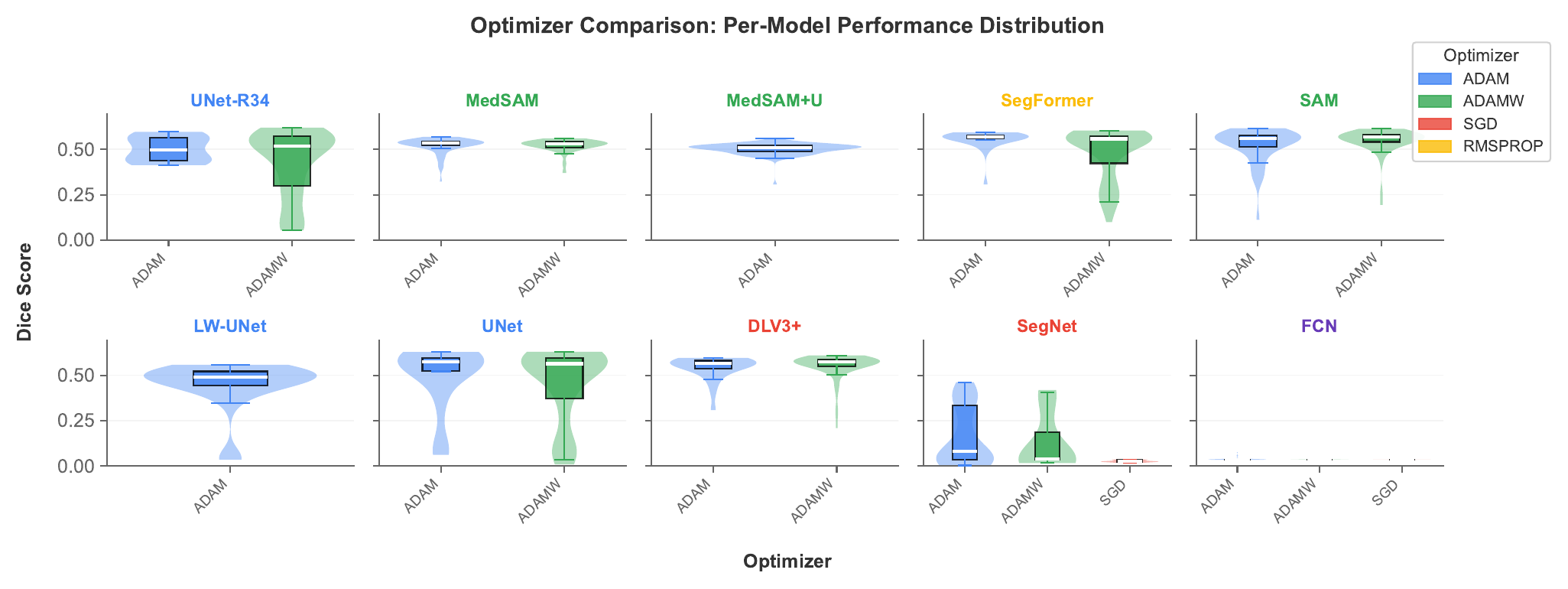}
\caption{Optimizer Comparison. Performance distributions across Adam, AdamW, SGD, and RMSprop for each architecture. AdamW generally outperforms alternatives for transformer-based models, while classical architectures show less optimizer sensitivity.}
\label{fig:hp_optimizer}
\end{figure}

To quantify how hyperparameter choices influence model performance, we analyzed 100-trial optimization sweeps for each architecture, examining learning rate, batch size, optimizer, scheduler, weight decay, and dropout. Figures \ref{fig:hp_learning_rate}, \ref{fig:hp_importance}, and \ref{fig:hp_optimizer} present this analysis, revealing architecture-specific sensitivities that further complicate the notion of fair model comparison.

Three findings emerge from this analysis. First, models exhibit substantially different optimal learning rate ranges. Foundation models (SAM, MedSAM) show narrow optimal ranges around $10^{-4}$, while classical architectures tolerate broader ranges, suggesting that under-tuned learning rates may systematically disadvantage certain architectures. Second, relative hyperparameter importance varies across architectures. Learning rate dominates for most models, but foundation models show greater sensitivity to weight decay (affecting fine-tuning behavior), while classical architectures are more sensitive to dropout and optimizer choice. Third, AdamW generally outperforms alternatives for transformer-based models (SegFormer, foundation models), while classical architectures show less optimizer sensitivity. This last finding suggests that optimizer choice, rarely reported in comparative studies, may introduce systematic bias.

Figure \ref{fig:hp_importance} demonstrates that the relative importance of each hyperparameter also varies by architecture beyond optimal ranges. Moreover, the choice of optimizer introduces an additional, often unreported source of systematic bias as illustrated in Figure \ref{fig:hp_optimizer}.

\subsubsection{Computational Efficiency}

Beyond segmentation accuracy, practical deployment requires considering computational costs. Table \ref{tab:efficiency} summarizes the efficiency metrics for all architectures, showing a wide range of resource requirements.

\begin{table}[h!]
\caption{Computational Efficiency Comparison. Metrics measured on NVIDIA V100 GPU with 256$\times$256 input resolution. Training time is the mean wall-clock time per LOOCV fold.}
\label{tab:efficiency}
\centering
\scriptsize
\begin{tabularx}{\textwidth}{@{} l *{7}{>{\centering\arraybackslash}X} @{}}
\toprule
\textbf{Model} & \textbf{Params (M) $\downarrow$} & \textbf{FLOPs (T) $\downarrow$} & \textbf{VRAM (GB) $\downarrow$} & \textbf{Latency (ms) $\downarrow$} & \textbf{FPS $\uparrow$} & \textbf{Train (s) $\downarrow$} & \textbf{LOOCV Dice $\uparrow$} \\
\midrule
SegFormer & \textbf{3.7} & \textbf{0.49} & \textbf{0.07} & 9.3 & 108 & 99 & 0.576 \\
Lightweight UNet & 7.8 & 1.02 & 0.08 & \textbf{2.7} & \textbf{365} & 86 & 0.249 \\
UNet-ResNet34 & 24.4 & 3.20 & 0.13 & 5.7 & 175 & 104 & 0.373 \\
DeepLabV3+ & 26.7 & 3.50 & 0.14 & 7.6 & 132 & 96 & 0.304 \\
SegNet & 29.5 & 3.86 & 0.20 & 5.7 & 174 & \textbf{71} & 0.105 \\
UNet & 31.4 & 4.11 & 0.25 & 7.7 & 131 & 145 & 0.392 \\
MedSAM & 95.8 & 12.56 & 2.71 & 121.8 & 8.2 & 259 & \textbf{0.694} \\
MedSAM+UNet & 95.7 & 12.55 & 2.71 & 137.3 & 7.3 & 173 & 0.479 \\
SAM & 97.3 & 12.75 & 2.72 & 120.9 & 8.3 & 311 & 0.496 \\
FCN & 134.4 & 17.61 & 5.40 & 14.3 & 70 & 90 & 0.121 \\
\bottomrule
\end{tabularx}
\end{table}

\begin{figure}[h!]
\centering
\includegraphics[width=0.9\textwidth]{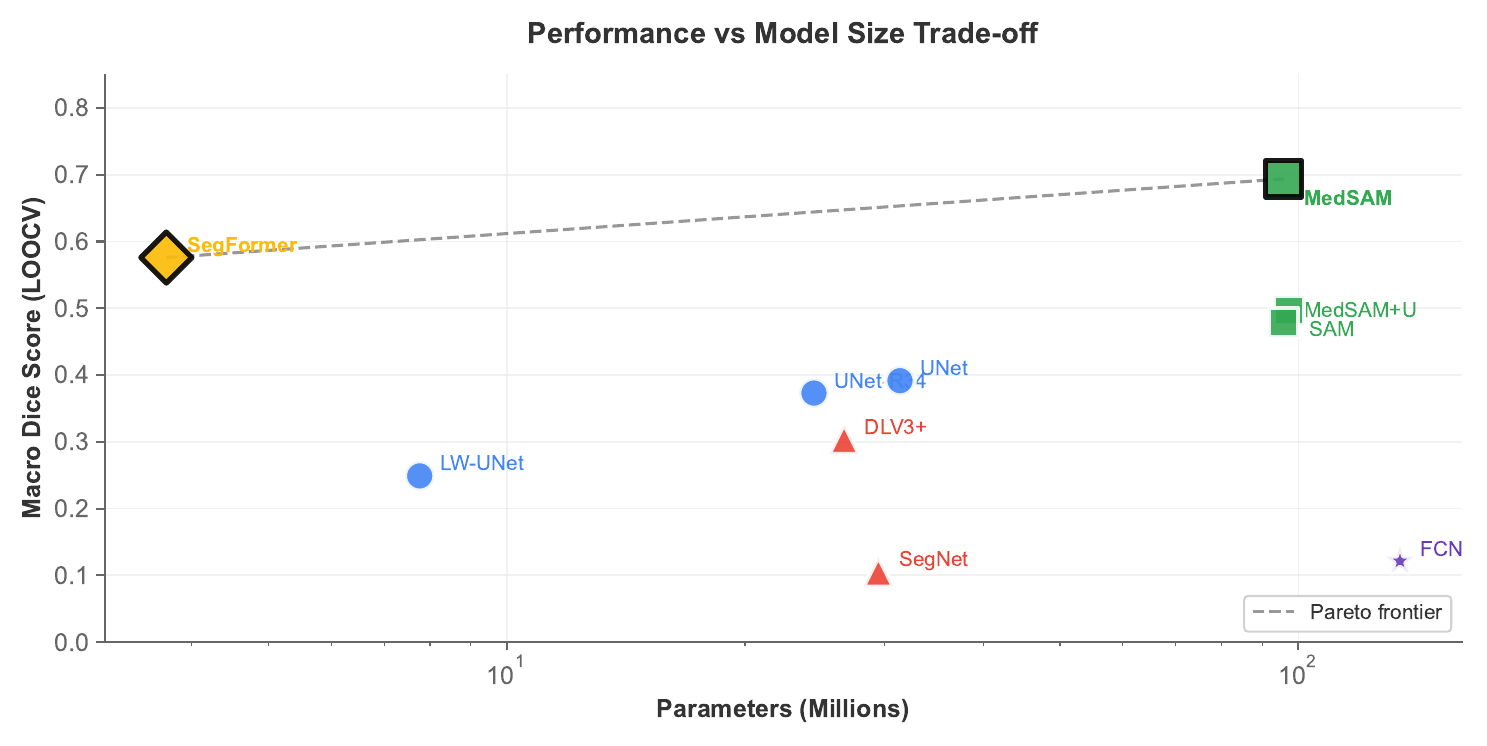}
\caption{Performance-Efficiency Trade-off. Bubble size represents model parameter count. SegFormer achieves the best balance of high performance with low computational cost, while foundation models (SAM, MedSAM) incur significant overhead.}
\label{fig:efficiency}
\end{figure}

The efficiency analysis reveals that \segformer{}, with only 3.7M parameters, achieves competitive performance with 25$\times$ fewer parameters than foundation models (Figure \ref{fig:efficiency}). This makes it suitable for resource-constrained deployment, despite slightly lower absolute performance.

\subsubsection{Boundary-Sensitive Metrics}\label{sec:boundary_metrics}

Beyond overlap-based metrics (Dice, IoU), accurate boundary delineation is important for histological analysis where tissue layer boundaries inform diagnosis. We therefore evaluated all models using boundary-sensitive metrics on the generalization dataset (DS2, $N=153$): the 95th percentile Hausdorff Distance (HD95) and Average Symmetric Surface Distance (ASSD) \cite{karimi2020reducing, reinke2024metrics}. We focus on DS2 boundary metrics as they reflect the more clinically relevant scenario of performance under distribution shift, where boundary accuracy differences between architectures are most pronounced.

Table \ref{tab:boundary_metrics} presents the boundary metric results. Lower values indicate better boundary accuracy. The results show a hierarchy: foundation models and transformers achieve the lowest boundary errors, with \segformer{} achieving HD95 of 27.2 pixels and ASSD of 8.3 pixels. Classical architectures (\fcn{}, \segnet{}) exhibit substantially larger boundary errors (HD95 $>$ 110 pixels), confirming their failure to learn precise tissue boundaries under distribution shift. Notably, the Dice scores under distribution shift are substantially lower than the in-distribution DS1 LOOCV values (e.g., \sam{} achieves 0.546 on DS2 vs. 0.496 on DS1, while \deeplab{} drops from 0.304 to 0.131), and the ranking order changes---\sam{} leads in generalization Dice despite ranking 3rd on DS1.

\begin{table}[h!]
\caption{Boundary-Sensitive Metrics on Generalization Dataset (DS2, $N=153$). HD95 = 95th percentile Hausdorff Distance; ASSD = Average Symmetric Surface Distance. Lower values indicate better boundary accuracy. Values are mean $\pm$ std across foreground classes (Lumen, Neointima, Media).}
\label{tab:boundary_metrics}
\centering
\scriptsize
\begin{tabularx}{\textwidth}{@{} l *{4}{>{\centering\arraybackslash}X} @{}}
\toprule
\textbf{Model} & \textbf{HD95 (px) $\downarrow$} & \textbf{ASSD (px) $\downarrow$} & \textbf{Dice $\uparrow$} & \textbf{Boundary Rank} \\
\midrule
\addlinespace[0.2em]
\multicolumn{5}{l}{\textit{Foundation Models \& Transformers}} \\
\addlinespace[0.1em]
\midrule
\textbf{SegFormer} & \textbf{27.2 $\pm$ 18.4} & \textbf{8.3 $\pm$ 6.6} & 0.502 & 1 \\
SAM & 47.9 $\pm$ 25.1 & 12.5 $\pm$ 7.7 & \textbf{0.546} & 2 \\
MedSAM & 33.1 $\pm$ 21.2 & 9.9 $\pm$ 7.8 & 0.488 & 3 \\
MedSAM+UNet & 38.9 $\pm$ 22.0 & 10.9 $\pm$ 7.6 & 0.477 & 4 \\
\midrule
\addlinespace[0.2em]
\multicolumn{5}{l}{\textit{Modern CNNs}} \\
\addlinespace[0.1em]
\midrule
UNet & 56.4 $\pm$ 24.9 & 16.8 $\pm$ 18.1 & 0.356 & 5 \\
Lightweight UNet & 56.7 $\pm$ 16.6 & 19.9 $\pm$ 6.8 & 0.333 & 6 \\
UNet-ResNet34 & 88.7 $\pm$ 11.4 & 32.6 $\pm$ 5.7 & 0.188 & 7 \\
DeepLabV3+ & 106.9 $\pm$ 13.8 & 54.0 $\pm$ 12.5 & 0.131 & 8 \\
\midrule
\addlinespace[0.2em]
\multicolumn{5}{l}{\textit{Classical Architectures}} \\
\addlinespace[0.1em]
\midrule
SegNet & 111.7 $\pm$ 8.8 & 50.1 $\pm$ 7.4 & 0.052 & 9 \\
FCN & 113.4 $\pm$ 8.0 & 53.0 $\pm$ 6.4 & 0.071 & 10 \\
\bottomrule
\end{tabularx}
\end{table}

\segformer{} achieves the best boundary metrics (HD95, ASSD) despite not having the highest generalization Dice score, suggesting that lightweight transformer architectures may learn more precise boundary representations than larger foundation models. The ranking inversion between boundary and overlap metrics---\sam{} leads in Dice while \segformer{} leads in boundary precision---reinforces our recommendation to consider multiple evaluation dimensions when selecting models for clinical deployment.

\FloatBarrier
\section{Discussion}\label{sec:discussion}


In this study, we comparatively evaluated ten different deep learning segmentation models on a small cardiovascular histology dataset to assess carotid artery segmentation for benchmarking. Our results based on extensive hyperparameter optimization and ablation studies showed that the concept of a single best model is unreliable and observed model rankings are largely influenced by evaluation protocols, statistical variability, and experimental settings. 
 
\subsection{Benchmark Reliability in Low-Data Settings}
The central finding of our work is that benchmark leadership is an artifact of the evaluation protocol, not an intrinsic property of a model. This instability is driven by a combination of the well-known bias-variance trade-off in validation strategies and the brittleness of hyperparameter optimization. As shown in our results in Figure~\ref{fig:loocv_ranking_traj_supp}, the high variance of LOOCV created unstable model rankings sensitive to single data points, while the higher bias of 3-Fold CV may have unfairly penalized more complex models.

This phenomenon provides strong empirical evidence for the illusion of control cognitive bias in machine learning research \cite{langer1975illusion}. The substantial effort invested in rigorous hyperparameter tuning creates overconfidence that randomness has been controlled and a definitive outcome has been reached. However, our findings show that even with optimal configurations, performance remains sensitive to the particular composition of data splits. This highlights a limitation in the standard practice of declaring a state-of-the-art model based on a single leaderboard, particularly when data is scarce. 

\subsection{Limitations of Standard Evaluation Metrics}
On in-distribution data, quantitative metrics suggest variability among
top-performing models. However, our multi-modal XAI analysis under distribution shift as shown in Figure~\ref{fig:qualitative_xai_comparison} reveals that these differences become more pronounced on unseen data: models that appeared statistically comparable on DS1 exhibit varying degrees of generalization on DS2, ranging from accurate predictions to substantial degradation. This suggests that in-distribution metric variability likely stems from minor, pixel-level boundary disagreements that may not be clinically meaningful \cite{maier2018rankings}, while generalization capacity represents a more informative criterion for distinguishing between
architectures.

\subsection{Generalization of Model Selection}
Our generalization experiments on DS2 (Section \ref{sec:generalization}) provide the most practically relevant insights. The ranking inversions observed under distribution shift---where models that performed well on DS1 failed on DS2---demonstrate that in-distribution benchmarking may not predict real-world deployment success. Foundation models (MedSAM, SAM) maintained reasonable performance under distribution shift, while classical architectures (FCN, SegNet) showed severe degradation.

Crucially, the in-distribution DS2 experiment (Section \ref{sec:ds2_indistribution}) demonstrates that these ranking inversions are not merely an artifact of distribution shift. When models are trained directly on DS2, the ranking hierarchy differs completely from DS1---\deeplab{} and \unet{} lead at $N=9$ on DS2, whereas \medsam{} led on DS1. Foundation models lose their advantage when in-distribution training data is available, confirming that rankings are dataset-specific, not just protocol-specific.

A methodological caveat is that our DS2 experiments reused hyperparameters optimized on DS1, which could systematically favor architectures whose optimal configurations transfer well across datasets. This design choice reflects a realistic deployment scenario, but the observed DS2 rankings may partly reflect hyperparameter transferability rather than pure architectural capacity.

This finding has direct implications for model selection in medical imaging. Rather than optimizing for marginal gains on a single benchmark, practitioners should prioritize robustness under distribution shift, where foundation models with diverse pre-training show superior generalization. Models should also be evaluated across varying sample sizes to assess stability. Furthermore, independent test sets from different institutions or imaging protocols are essential for domain-specific validation. Finally, rankings established on one dataset should not be assumed to transfer to another, even at matching sample sizes, underscoring the need for cross-dataset validation.

\subsection{The Statistical Stability of Performance Rankings}
This study was conducted in a very low-data environment ($N=9$), which serves as a challenging test case for benchmarking practices. Our sample size sensitivity analysis in Figure \ref{fig:ds2_rank_trajectory}a--b empirically maps the phase transition from instability to stability, with the rank correlation convergence in Figure \ref{fig:ds2_rank_trajectory}e and rank swap analysis in Figure \ref{fig:ds2_rank_trajectory}f confirming that rankings become reproducible only at $N \geq 50$ samples.

The volatile model rankings at small sample sizes, particularly under the high-variance LOOCV protocol as shown in Figure~\ref{fig:loocv_ranking_traj_supp}, are an expected consequence of high variance estimators. However, our ablation studies (Section \ref{sec:ablation}) reveal that variance arises from multiple sources---not just data splits, but also augmentation choices, random seed, and hyperparameter selection. Even with controlled randomness, the small dataset size introduces substantial variance.

Importantly, the DS2 in-distribution experiment (Section \ref{sec:ds2_indistribution}) independently confirms this stability threshold: rankings on DS2 are volatile at $N \leq 25$ but stabilize at $N \geq 50$, providing cross-dataset validation of this finding.

We therefore propose evidence-based guidelines: confident model selection requires a minimum of 50--100 samples for stable rankings, and performance claims should always be accompanied by bootstrap confidence intervals and cross-protocol validation. This guidance is essential for researchers in rare diseases or other data-limited domains.

\subsection{Understanding Hyperparameter Sensitivity Across Architectures}
Our hyperparameter optimization experiments revealed architecture-specific sensitivities that merit mechanistic explanation, as these insights are crucial for practitioners seeking to replicate our findings or adapt these models to new domains.

Learning Rate Preferences: The optimal learning rate varies substantially across architecture families. Transformer-based models (\segformer{}, MiT-B0 encoder) converged optimally with learning rates around $6 \times 10^{-5}$, consistent with established fine-tuning guidelines for pre-trained vision transformers \cite{huggingface2023segformer}. This conservative learning rate preserves the rich representations learned during pre-training while allowing task-specific adaptation. In contrast, foundation models (SAM, MedSAM) using parameter-efficient fine-tuning (Norm Tuning) favored slightly higher learning rates ($1 \times 10^{-4}$ to $3 \times 10^{-4}$), as the limited number of trainable normalization parameters require sufficient gradient magnitude to learn meaningful domain adaptations without disrupting the frozen backbone \cite{mazurowski2024finetune}. Classical CNNs trained from scratch exhibited broader tolerance, performing reasonably across a wider learning rate range.

Loss Function Selection: The dominance of Focal-Dice loss across all architectures (Table 2) is explained by its dual mechanism for handling the class imbalance inherent in our dataset, where tissue layers occupy substantially different pixel proportions. The Dice component provides spatial overlap sensitivity that is inherently robust to class imbalance, while the Focal component down-weights easy examples to focus learning on difficult boundary pixels \cite{yeung2022unified}. This combination addresses both class imbalance (different tissue proportions) and difficulty imbalance (easy interior vs. hard boundary pixels), making it particularly effective for medical image segmentation where accurate boundary delineation is clinically critical.

Optimizer Choice: AdamW consistently outperformed alternatives for transformer-based architectures, likely due to its decoupled weight decay regularization which prevents overfitting in over-parameterized models \cite{loshchilov2019adamw}. Classical CNNs showed less optimizer sensitivity, with Adam and SGD with momentum producing comparable results. These findings underscore that optimizer choice can systematically bias results and should be tuned independently for each architecture family.


\section{Conclusion}\label{sec:conclusion}

In this study, we conducted a comprehensive evaluation of
ten deep learning segmentation models on a realistic, data-scarce medical dataset of cardiovascular histology images, to assess their performance in segmenting carotid artery structures. Based on the evaluations through extensive hyperparameter optimization runs and ablation experiments, we empirically observed that model rankings depend on the evaluation protocol, statistical noise, and experimental conditions rather than reflecting true architectural advantage.

Our findings indicate that bootstrap confidence intervals for top-performing models substantially overlap, indicating statistical indistinguishability; ablation studies reveal that augmentation choices and random seeds introduce variance comparable to architectural differences. Furthermore, generalization experiments on an independent dataset showed substantial ranking inversions under distribution shift, with foundation models maintaining performance while classical architectures fail. Moreover, sample size sensitivity analysis showed that stable rankings require at least 50-100 samples; and within-distribution training on varying sample sizes reveals that model ranking hierarchies are dataset-specific.

\section*{Author contributions: CRediT}

\textbf{PMK:} Conceptualization, Data curation, Formal analysis, Methodology, Software, Visualization, Validation, Writing – original draft;
\textbf{AAP:} Conceptualization, Data curation, Methodology, Validation, Writing – review and editing; 
\textbf{YS:} Conceptualization, Methodology, Validation, Writing – review and editing; 
\textbf{ZL:} Conceptualization, Methodology, Validation, Writing – review and editing;
\textbf{MT:} Data curation, Funding acquisition, Resources, Validation, Writing – review and editing;
\textbf{AC:} Investigation, Data curation, Resources, Validation, Writing – review and editing;
\textbf{EAL:} Conceptualization, Resources, Funding acquisition, Providing samples, Project administration, Supervision, Writing – review and editing; 
\textbf{SA:} Conceptualization, Funding acquisition, Methodology, Validation, Supervision, Writing – review and editing.
\newline

\section*{Declaration of competing interest}

The authors declare that they have no known competing financial interests or personal relationships that could have appeared to influence the work reported in this paper.

\section*{Acknowledgements}

The authors gratefully acknowledge the contributions of Brijesh Kumar Singh, and Priyanka Gupta (Duke-NUS Medical School, Singapore), Roshni R. Singaraja (National University of Singapore), and Elisa Octavia Velicu (University of Medicine and Pharmacy Carol Davila, Bucharest) for their scientific and technical support in this study.

\section*{Funding}

This work is supported by the Innovation Fund Denmark for the project DIREC (9142-00001B), CNCS—UEFISCDI, project number PN-III-P4-PCE-2021-1680. 
The content is solely the responsibility of the authors and does not represent the official views of funding sources.

\section*{Data availability}

Data will be made available on request.

\bibliographystyle{elsarticle-num} 
\bibliography{references}

\newpage
\appendix

\section*{Supplementary Materials}
\renewcommand{\thetable}{S\arabic{table}}
\renewcommand{\thefigure}{S\arabic{figure}}
\renewcommand{\theequation}{S\arabic{equation}}

\section{Bayesian Hyperparameter Optimization Details}\label{sec:supp_bayesian}
Our hyperparameter optimization uses the Tree-structured Parzen Estimator (TPE) \cite{bergstra2011tpe}, implemented via Optuna with a fixed random seed ($seed = 42$) and 10 startup trials of random sampling before switching to the TPE sampler.

\textbf{Why TPE, not Gaussian Processes.}
Classical Bayesian optimization fits a Gaussian Process (GP) to model the objective $f(x)$ directly, then maximizes an acquisition function (e.g., Expected Improvement) over that surrogate. This requires inverting a covariance matrix at cost $\mathcal{O}(t^3)$ per step, which becomes prohibitive as the number of trials $t$ grows. TPE avoids this entirely by modeling the \emph{inverse} relationship: instead of $p(y \mid x)$, it models $p(x \mid y)$.

\textbf{Observation splitting.}
After $t$ evaluations, TPE partitions the observed hyperparameter--score pairs at a quantile threshold $\gamma$. Configurations with scores better than the threshold $y^*$ define a ``good'' density $\ell(x)$; the remainder define a ``bad'' density $g(x)$:
\begin{equation}
\label{eq:tpe_densities}
p(x \mid y) = \begin{cases}
\ell(x) & \text{if } y < y^* \\
g(x) & \text{if } y \geq y^*
\end{cases}
\end{equation}
where $y^*$ is the $\gamma$-quantile of observed scores ($\gamma = 0.25$ in Optuna) and both $\ell(x)$ and $g(x)$ are estimated via Parzen window (kernel density) estimators.

\textbf{Selection criterion.}
Bergstra et al.\ \cite{bergstra2011tpe} show that, under this formulation, the next configuration should be chosen to maximize the ratio of the two densities:
\begin{equation}
\label{eq:tpe_ratio}
x_{\text{next}} = \arg\max_x \; \frac{\ell(x)}{g(x)}
\end{equation}
Intuitively, this selects hyperparameters that are \textit{likely under good trials} and \textit{unlikely under bad trials}. In practice, TPE draws many candidates from $\ell(x)$ and picks the one with the highest $\ell(x)/g(x)$ ratio. This naturally balances exploitation (sampling where good configurations cluster) with exploration (avoiding regions dominated by poor configurations), enabling efficient navigation of high-dimensional hyperparameter spaces.

\section{Hardware and Software Specifications}\label{sec:supp_hardware}
For complete reproducibility, we provide detailed specifications of our computational environment in Table \ref{tab:supp_hardware}.

\begin{table}[p]
\centering
\caption{Hardware and Software Environment Specifications}
\label{tab:supp_hardware}
\footnotesize
\begin{tabularx}{\textwidth}{@{} l X @{}}
\toprule
\textbf{Component} & \textbf{Specification} \\
\midrule
\addlinespace[0.3em]
\multicolumn{2}{l}{\textit{\textbf{Hardware (Main Experiments)}}} \\
\addlinespace[0.2em]
\midrule
GPU & 1--3$\times$ NVIDIA Tesla V100 (32GB VRAM each) \\
GPU Architecture & Volta \\
CUDA Cores & 5,120 per GPU \\
Memory Bandwidth & 900 GB/s \\
\midrule
\addlinespace[0.3em]
\multicolumn{2}{l}{\textit{\textbf{Hardware (Resolution Ablation)}}} \\
\addlinespace[0.2em]
\midrule
GPU & 3$\times$ NVIDIA Tesla V100-SXM2 (32GB VRAM each) \\
GPU Architecture & Volta \\
Provider & UCloud (\url{https://cloud.sdu.dk}) \\
\midrule
\addlinespace[0.3em]
\multicolumn{2}{l}{\textit{\textbf{Software Environment}}} \\
\addlinespace[0.2em]
\midrule
Deep Learning Framework & PyTorch 2.1+ \\
CUDA Version & 12.1 \\
cuDNN Version & 8.9 \\
Python Version & 3.10+ \\
Segmentation Library & segmentation-models-pytorch 0.3.3+ \\
Foundation Models & transformers 4.36+ \\
Hyperparameter Optimization & Optuna 3.4+ \\
Experiment Tracking & JSON-based offline storage \\
\midrule
\addlinespace[0.3em]
\multicolumn{2}{l}{\textit{\textbf{Reproducibility Settings}}} \\
\addlinespace[0.2em]
\midrule
Random Seed & 42 \\
Python Random & Seeded \\
NumPy Random & Seeded \\
PyTorch Manual Seed & Seeded \\
CUDA Manual Seed & Seeded (all devices) \\
cuDNN Deterministic & True \\
cuDNN Benchmark & False \\
Mixed Precision (AMP) & Enabled \\
\midrule
\addlinespace[0.3em]
\multicolumn{2}{l}{\textit{\textbf{Training Configuration}}} \\
\addlinespace[0.2em]
\midrule
Data Workers & 8 \\
Pin Memory & Enabled \\
Early Stopping Patience & 20 epochs \\
Maximum Epochs & 200 \\
\bottomrule
\end{tabularx}
\end{table}

Each individual experiment was executed on a single NVIDIA Tesla V100 GPU to eliminate hardware-induced variance; up to three identical GPUs were used concurrently to run different model experiments in parallel. The deterministic settings ensure that given the same random seed, data splits, and hyperparameters, our results can be practically reproduced. We note that Automatic Mixed Precision (AMP) may introduce minor non-determinism in floating-point reduction operations; however, this has negligible impact on aggregate metrics and conclusions. 

\end{document}